\documentclass[10pt,twocolumn,letterpaper]{article}

\usepackage{wacv}
\usepackage{times}
\usepackage{epsfig}
\usepackage{graphicx}
\usepackage{amsmath}
\usepackage{amssymb}
\usepackage{multirow}
\usepackage[algo2e]{algorithm2e}
\usepackage{setspace}
\usepackage{algorithm}
\usepackage[noend]{algpseudocode}
\usepackage{scalerel}
\usepackage{subcaption}
\makeatletter
\newcommand{\appendixhead}%
{\centering {\huge Supplementary}
\vspace{0.25in}}
\makeatother
\def\wacvPaperID{453} % Enter the WACV Paper ID here

\wacvfinalcopy % *** Uncomment this line for the final submission

\ifwacvfinal
\fi

\ifwacvfinal
\usepackage[breaklinks=true,bookmarks=false]{hyperref}
\else
\usepackage[pagebackref=true,breaklinks=true,colorlinks,bookmarks=false]{hyperref}
\fi

\begin{document}

%%%%%%%%% TITLE
% \title{LT-GAN: Latent transformation detection as Self-Supervised GAN for Steerable Latent Space}

\title{LT-GAN: Self-Supervised GAN with Latent Transformation Detection}
\author{Parth Patel \thanks{Authors contributed equally}  \thanks{Work done during Adobe MDSR internship}  $^{2}$ \\
{\tt\small f2016150p@alumni.bits-pilani.ac.in}
\and
Nupur Kumari$^{*1}$ \\
{\tt\small nupkumar@adobe.com}
\and
Mayank Singh$^{*1}$ \\
{\tt\small msingh@adobe.com}
\and
Balaji Krishnamurthy$^{1}$ \\
% Adobe Inc, Noida, India \\
{\tt\small kbalaji@adobe.com} \\
1. Media and Data Science Research Lab, Adobe \\
2. Birla Institute of Technology \& Science, Pilani India 
}
\maketitle
\ifwacvfinal\thispagestyle{empty}\fi

\maketitle
%\thispagestyle{empty}

%%%%%%%%% ABSTRACT
\begin{abstract}
Generative Adversarial Networks (GANs) coupled with self-supervised tasks, have shown promising results in unconditional and semi-supervised image generation. We propose a self-supervised approach (LT-GAN) to improve the generation quality and diversity of images by estimating the GAN-induced transformation (i.e. transformation induced in the generated images by perturbing the latent space of generator). Specifically, given two pairs of images where each pair comprises of a generated image and its transformed version, the self-supervision task aims to identify whether the latent transformation applied in the given pair is same to that of the other pair. Hence, this auxiliary loss encourages the generator to produce images that are distinguishable by the auxiliary network, which in turn promotes the synthesis of semantically consistent images with respect to latent transformations. We show the efficacy of this pretext task by improving the image generation quality in terms of FID on state-of-the-art models for both conditional and unconditional settings on CIFAR-10, CelebA-HQ and ImageNet datasets. Moreover, we empirically show that LT-GAN helps in improving controlled image editing for CelebA-HQ and ImageNet over baseline models. We experimentally demonstrate that our proposed LT self-supervision task can be effectively combined with other state-of-the-art training techniques for added benefits. Consequently, we show that our approach achieves the new state-of-the-art FID score of 9.8 on conditional CIFAR-10 image generation.

\end{abstract}

%%%%%%%%% BODY TEXT
\section{Introduction}
% Para 1 : Introduce GANs and usefulness of semantic editing
Generative Adversarial Networks (GANs) have become a popular class of generative models as they have shown impressive capacity in modelling complex data distributions, such as images \cite{biggan2018brock,stylegan2019karras} and audio \cite{audio_GAN,audio_gan2}. GANs consist of a generator and a discriminator network with competing goals: the generator's objective is to generate realistic samples to fool the discriminator and the discriminator's objective is to distinguish between the real samples and the fake ones synthesized by the generator. The generator learns a mapping from a latent distribution to the data distribution via adversarial training with the discriminator. Despite the significant progress of GANs, there lacks enough understanding of how different semantics are encoded in the latent space of generator. It has been observed that in a well trained GAN, semantics encoded in the latent space are disentangled to some extent which makes it possible to perform controlled image editing \cite{interface2020shen,gansteerability, controlling2020iclr}. 

% The core idea foris that a semantic trait of interest, can be modified independently of other traits. Although the definitions of disentanglement representations are varied,

% Para 2 : Using the Motivation of disentangled space introduce your intuition for ss Task interfaceGAN

% Para 3 : Introduce some SS loss earlier used in GANs (Rotnet, FXNet) and their scope in general 
Another class of unsupervised learning called self-supervision has demonstrated promising results on a diverse set of computer vision tasks \cite{s2m22020, Spyros2018rotate, arrowtime2019abhinav, color2018,downstream2019abhinav}. This training paradigm usually involves designing an auxiliary task with pseudo-labels derived from the structural information of the data.
% to learn better feature representation useful for downstream applications \cite{downstream2019abhinav}.
Self-supervised learning has also been used in collaboration with adversarial training in GANs \cite{rot_ssgan2019chen, fxgan2020wacv} to improve training stability and unconditional/semi-supervised image generation quality \cite{semisupervised2019icml}. Generally, the role of self-supervised methods in GANs is to regularize the discriminator which in turn guides the generator to produce images with more informed geometric or global structure. For example, SS-GAN \cite{rot_ssgan2019chen} introduced an auxiliary task of predicting the degree of rotation in the input image to the discriminator during GAN training.
% This lead to better feature representation learning by utilising the information about geometric transformation of images.

% Para 4 : Briefly discuss the task of SS ie pair wise concatenation to distinguish different changes in images.

% In contrast to the conventional unsupervised feature learning paradigm of auto encoding the input data \cite{autoencoder2008icml,autoencoderv1994},
The authors of \cite{aet2019transformation} propose to learn the unsupervised feature representation by encoding the input data transformation rather than data itself \cite{autoencoder2008icml,autoencoderv1994}. Specifically, in AET \cite{aet2019transformation}, image transformation operators are sampled, and the objective is to estimate the transformation given the feature representation of the original and the transformed images. This framework of unsupervised feature learning encourages encoding of the essential visual structure of the transformation so that the transformation can be predicted. \cite{aet2019transformation} has shown promising results on various standard unsupervised visual downstream tasks. Inspired by this approach \cite{aet2019transformation}, in the domain of GAN, we propose a binary self-supervised task that aims to detect if the GAN-induced transformation (as described in \cite{aet2019transformation}) applied on two generated images is same. The key idea in this transformation prediction task is to promote useful representation learning as the features would have to encode sufficient information about visual structures of both the original and transformed images for auxiliary training. Note that this is in contrast with the previous line of works\cite{crgan2019chen} that augments the input data to the discriminator using a fixed set of static transformations and penalizes the sensitivity of discriminatory features to these transformations.

% Display Page Nos.

In this work, we propose a self-supervised task called Latent Transformation (LT) for improving the quality of image synthesis and latent space semantics. Previous methods make use of limited predetermined augmentation or transformations (such as rotation, translation) to define the self-supervised loss. However, we utilize GAN-induced transformations as described in \cite{aet2019transformation} to define our pretext task. In contrast to earlier works \cite{ssganadv2019,fxgan2020wacv} that adds a self-supervised loss to the discriminator, our auxiliary task promotes the generator to synthesize images such that the GAN-induced transformations are distinguishable at the feature representation level. 

% 
% Para 5 : Summarize your contributions in bullets
Our main contributions in this paper are the following :
\begin{itemize}

\item We propose a novel self-supervised task of identifying GAN induced latent transformations to optimize the generator in collaboration with the adversarial training of GANs.

\item We demonstrate the efficacy of our proposed LT-GAN approach on several standard datasets with various architectures by improving the conditional and unconditional state-of-the-art image generation performance on Fr\'echet inception distance (FID)\cite{fid2017martin} metric.
% both inception score (IS) and 

\item We empirically show that our LT-GAN model improves controlled image editing using existing semantic editing frameworks \cite{interface2020shen,controlling2020iclr} over baseline models.
% the performance on semantic image editing methods using latent space manipulation, highlighting the disentangled latent space feature learning.

% through latent space manipulation

\end{itemize}

\section{Background and Related Work}

\paragraph{Self-Supervised learning} is an unsupervised learning framework that seeks to leverage supervisory signals from the structural information present in the data by defining auxiliary pretext tasks. Self-supervision techniques have shown a huge potential in a diverse set of research tasks, ranging from robotics to computer vision \cite{robot2018,crossmodal2018,multisensory2018,s2m22020,Spyros2018rotate}. In visual domain, a pretext task is designed with labels derived from the images themselves that help in learning rich feature representation useful for downstream tasks \cite{downstream2019abhinav}. Some of the earliest effort \cite{doersch2015seminal} in this area utilize relative position prediction of image patches. Inspired by this task's relation to prediction of context in images, the authors of \cite{Noroozi2018jigsaw,Nathan2018improve,Noroozi2018improve} use a pretext task of predicting the permutation in a image with shuffled patches. \cite{Spyros2018rotate} used the surrogate objective of predicting the angle of rotation for unlabelled image. The task of in-painting \cite{painting2018} and image colorization \cite{color2018,color_2018_1} have also been used as auxiliary tasks in the self-supervised learning framework. In contrast with utilizing the geometric and structural in-variances, \cite{cluster2018self} uses the task of predicting the cluster assignment in feature space as pseudo labels for unlabeled data. 
% This works by alternating between clustering of the image descriptors and updating the network by predicting the cluster assignments. 
Also, \cite{Noroozi2017count} obtains supervision signal by counting the visual primitives present in the patches of images. Along the lines of transformation prediction task like \cite{Spyros2018rotate}, AET\cite{aet2019transformation} introduces a surrogate task of reconstruction of input data transformations to learn unsupervised feature representations. Inspired by this work, our approach LT-GAN proposes the auxiliary task of estimating GAN-induced transformations. We hypothesize that it would encourage the generator to synthesize semantically consistent image transformations with respect to similar latent space perturbations. 

% to focus on capturing the latent space perturbations while generating images.

%  the change in original and transformed images are semantically consistent across all generated images (e.g translation, background change).
 
% the discriminatory latent transformation features.

\paragraph{GANs with self-supervised auxiliary tasks}
Recently, self-supervised learning has been coupled with adversarial training to improve the training stability and image quality of GANs \cite{fxgan2020wacv,rot_ssgan2019chen, tran_2019_neurips_gan}. The motivation behind adding self-supervised loss to GAN training is to equip the feature representations to recognize the global structures present in real data through the pretext tasks. SS-GAN \cite{rot_ssgan2019chen} uses the auxiliary task of image rotation degree classification based on the discriminator features. The authors of \cite{fxgan2020wacv} propose to use the pretext task of distinguishing between real images and corrupted real images with GAN training. These corrupted images are created by randomly exchanging pairs of patches in an image’s convolutional feature map. 
% The existing self-supervised GAN approaches optimize the discriminator using an auxiliary objective, however in our approach LT-GAN, we promote the generator for improved latent representation learning.

\paragraph{Latent Space Manipulation for semantic editing in GANs}
% Image editing through latent space manipulation in GANs is usually achieved by optimizing for a linear direction which when added to the latent vector results in target image \cite{}.  A related field of research is learning disentangled latent space structure. 
Conventional approaches of finding interpretable manipulations in GAN latent space compute linear directions corresponding to attribute change by using annotated attributes tags of the images \cite{dcgan2016radford, stylegan2019karras}. \cite{dfi2017upchurch} showed this to be even true for pre-trained classifier where interpolation in latent feature space of target and source images leads to interpretable transfer of visual properties from a source image to a target image. To assume control over the image generation process in GANs, work by \cite{condgan2017odena,infogan2016} propose modifications in architecture and training approach. \cite{condgan2017odena} allows the generation of images belonging to a certain class and therefore requires access to labels for training the model. \cite{infogan2016} learns disentangled representations by maximizing the mutual information between a subset of the latent variables and the observation which enables the process of finding a posteriori semantic direction. However, work by \cite{gansteerability, controlling2020iclr, ganalyze2019isola} shows that the latent space directions corresponding to transformations (such as zoom, scale, shift, brightness) can be computed using the respective augmentation of images on pre-trained GAN models. These approaches \cite{gansteerability, controlling2020iclr} lighten the requirement of attribute tagged images for some general image editing tasks and can also serves as a measure for generalization capacity of generative models. The performance of these latent self-supervised trajectories are limited by biases in the training dataset and the model's generalization performance \cite{denton2019bias, gansteerability}. Recent advances in GANs \cite{biggan2018brock,stylegan2019karras} in generating photo-realistic images have unlocked the potential for content creation and fine tuning modifications \cite{zhu2016finetune,zhu2019finetune}. \cite{interface2020shen} performs semantic face editing (for changing attributes such as age, expression, etc.) on a fixed pre-trained GAN model by using linear subspace projection techniques and thus demonstrating disentanglement of the latent space of pre-trained GANs. We show that our approach LT-GAN improves controlled image editing over baseline models by using existing semantic editing frameworks \cite{interface2020shen,controlling2020iclr}.

% The correlation between directions corresponding to manipulation of different semantic attributes can provide us with a metric to compute the degree of semnatic disentanglement in the latent space of GANs. 

% \paragraph{Evaluation techniques for disentangled learning}

\section{Methodology}
In this section, we first present the standard GAN formulation and terminologies used in the paper. We then introduce our training methodology for LT-GAN that leverages a self-supervised task defined on the latent space of generator to better organize the semantics encoded in the latent space and promote diverse image generation.
% to make it more steerable. 

\begin{figure*}[t]
\centering
\scalebox{0.9}{
\centering
   \includegraphics[width=\linewidth]{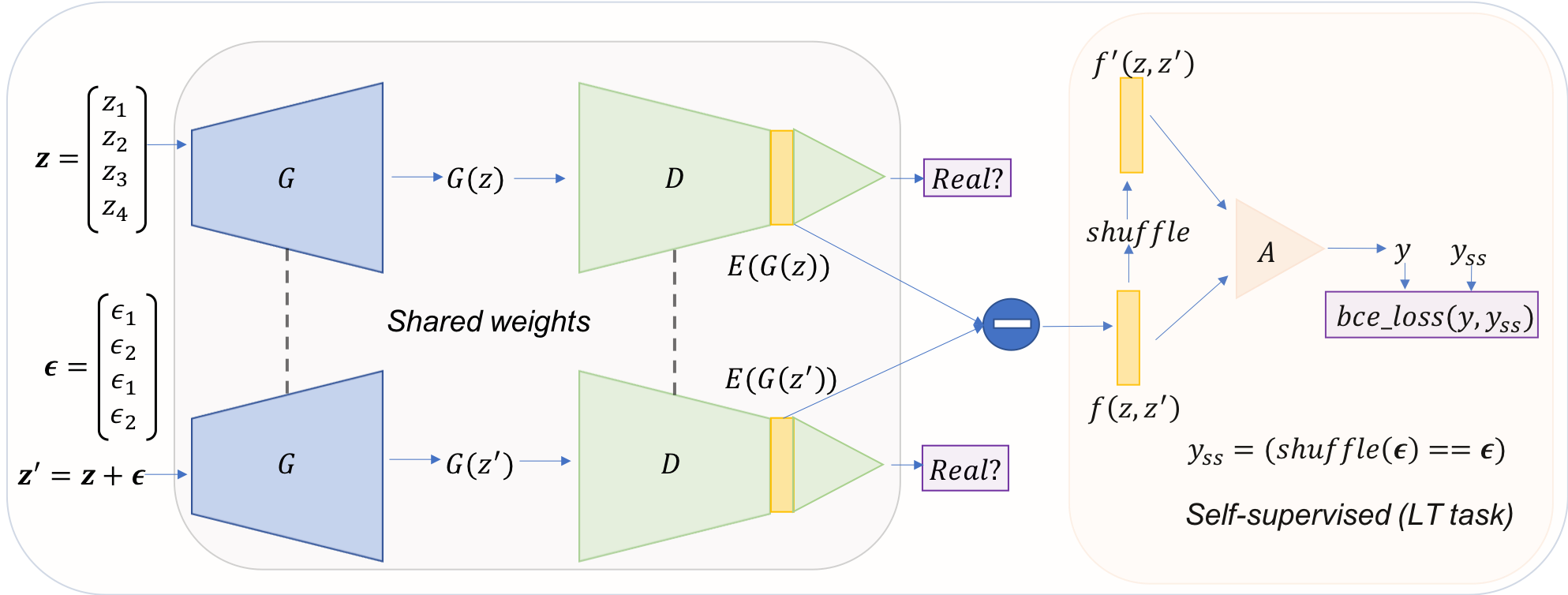}
   }
   \caption{\footnotesize{Overview of our proposed LT-GAN self-supervision task for generator training. Generated images $G(z)$ and its GAN-induced transformation $G(z+\epsilon)$ are used for defining the self-supervision loss ($bce\_loss$). Given intermediate discriminator features of above generated images i.e. $E(G(z))$ and $E(G(z+\epsilon))$, the feature representation of the GAN-induced transformation is $f(z , z+\epsilon)$. Auxiliary network $A$ and generator $G$ are trained simultaneously on pretext task to predict if $f(z_1,z_1+\epsilon_1)$ and  $f(z_2,z_2+\epsilon_2)$ features are generated from same $\epsilon$ perturbation in the latent space. }}
\label{fig:flowchart}
\end{figure*}

\subsection{Generative Adversarial Networks} 
Generative adversarial network (GAN) consists of a generator $G:z \rightarrow x $ and a discriminator $D:x \rightarrow \mathbb{R}$. $G$ learns a mapping from the latent code $z \in \mathbb{R}^d $ sampled from a prior distribution $p(z)$ to an observation $x \in \mathbb{R}^n$ e.g. a natural image manifold. The role of discriminator $D$ is to differentiate between samples from real data distribution $p(x)$ and the ones generated from $G$. The standard training of GAN involves minimizing the following loss function in an alternating fashion:
\begin{equation}
\begin{aligned}
    L_D &: - \mathbb{E}_{x  \sim p(x) } [\log(D(x))] - \mathbb{E}_{x \sim p(z) } [1- \log(D(G(z))) ] \\
    L_G &: - \mathbb{E}_{z \sim p(z) } [\log(D(G(z))) ]
\end{aligned}
\end{equation}
This loss function is commonly known as non-saturating loss and was originally proposed in \cite{gan_goodfellow}. A notable modification of the loss for improved training is the hinge loss \cite{hinge_loss_gan}: 
\begin{equation}
\begin{aligned}
    &L_D : \mathbb{E}_{x  \sim p(x) } [1 - D(x) ]_{+} + \mathbb{E}_{x \sim p(z) } [1 + D(G(z)) ]_{+} \\
    &L_G : - \mathbb{E}_{z \sim p(z) } [D(G(z)) ] \\
    &\text{where } [y]_{+} = max(0 , y)
\end{aligned}\label{eq:hinge-loss}
\end{equation}

Here, latent code $z$ is usually sampled from a normal distribution and for each step of generator update, discriminator is updated for $d_{step}$ times. A common issue with GANs is its instability during training that generally requires stabilization techniques \cite{inception2016,wgan_improved,sngan_proj}. In this work, we use the widely accepted practice of spectral normalization \cite{sngan_proj,biggan2018brock} for stable training. 

% Traditional approaches learn the latent space trajectory by optimizing for metrics like perceptual loss \cite{perceptual_metric} between the generated and target images \cite{gansteerability, controlling2020iclr,interface2020shen}. 
% A desirable property that enables precise control for manipulating target attributes of generated image is a disentangled latent space with different semantic aspects embedded in linear subspaces. 
% The auxiliary task promotes generation of similarly transformed images when latent codes are manipulated in a common direction in their local neighbourhood.  
% which has recently received significant attention 
\subsection{Latent Transformation GAN (LT-GAN)}
One of the attractive use-cases of GAN which has recently received significant attention is controlled image synthesis by latent space manipulation. We introduce a self-supervision task on the generator for improved steerability in latent space by leveraging GAN-induced transformations. 

\vspace{-11pt}
\paragraph{GAN-induced Transformation}
Give a latent code $z$ sampled from a prior distribution $p(z)$ and the corresponding generated image $I = G(z)$, we define GAN-induced transformation as:
\begin{equation}
    \mathcal{T}_{\epsilon}(G(z)) = G(z+\epsilon) \; : \;\epsilon \sim p(\epsilon)
\end{equation}

For a fixed generator, the transformation $\mathcal{T}$ is parametrized by $\epsilon$, a perturbation of small magnitude, sampled from a distribution $p(\epsilon)$. Applying $\mathcal{T_{\epsilon}}$ to the image $I$ generated from latent code $z$ generates a transformed version of the image $\mathcal{T_{\epsilon}}(I)$. In our self-supervision task, we aim to enforce that when a transformation $\mathcal{T}$ parametrized by an $\epsilon$ is applied to latent codes, the change in original and transformed images are semantically consistent across all generated images (e.g translation, background change).

% Our goal in this paper is to train the generator in a self-supervised manner such that the encoding of semantics in the latent space is disentangled i.e change in latent code in a specific direction leads to similar transformation in the generated images. To achieve this, we define an auxiliary task to identify similarity of transformations between a pair of generated images using discriminator features of the image.

% \noindent For two latent vectors $z_1$, $z_2$ and $\epsilon_1$, a small magnitude noise vector, $ I_{z_1} = G(z_1)$ and $I_{z_1 + \epsilon_1} = G(z_1 + \epsilon_1)$ are images with slight variations between them, similarly $I_{z_2}= G(z_2)$ and $I_{z_2 + \epsilon_1} = G(z_2 + \epsilon_1)$ are slight variations of each other. We consider $I_{z_1+ \epsilon_1}$ and $I_{z_2+ \epsilon_1}$ as a similar generative transformation of images $I_{z_1}$ and $I_{z_2}$ (e.g. translation ) when the latent code is perturbed in a specific direction $\epsilon_1$ in its local neighbourhood. Similarly, perturbing the latent code in a different direction $\epsilon_2$ leads to a different transformation in the generated image. We can quantify this generative transformation $t_{\epsilon}$ between two images $G(z)$ and $G(z+\epsilon)$ as the difference in feature representation of the images. 

\vspace{-8pt}
\paragraph{LT Self Supervision} 
Let, $E: x \rightarrow E(x)$ be an encoder network to extract the features of an image. Given a transformation $\mathcal{T_{\epsilon}}$, the feature representation corresponding to the change between original and transformed image can be written as:
\begin{equation}
f(z,z+\epsilon) = E(G(z)) -  E((\mathcal{T_{\epsilon}}(G(z))
    % f:[E(G(z)),E((\mathcal{T_{\epsilon}}(G(z))] \rightarrow f(z , z + \epsilon) 
\end{equation}\label{eq:change_in_t}
Where $f$ captures the change in the image feature and its GAN-induced transformation. We choose to implement it simply as the subtraction of encoder features, though other operations like concatenation are valid choices to explore. In LT self supervision, given $f_1 = f(z_1, z_1 + \epsilon_1)$ and $f_2 = f(z_2, z_2 + \epsilon_2)$ where $z_1 , z_2 \sim p(z)$, we introduce an auxiliary network $A$ to classify whether the above pair of features $\{f_1,f_2\}$ are corresponding to transformations parameterized by same $\epsilon$ or different. Specifically, the self-supervision loss is defined as: 
\begin{equation}
\begin{aligned}
        & L_{A} = \mathop{\mathbb{E}}_{\substack{z_1 , z_2 \sim p(z) \\ \epsilon_1,\epsilon_2 \sim p(\epsilon)}} L \Big( A\big([ f(z_1, z_1 + \epsilon_1) , f(z_2, z_2 + \epsilon_2) ]\big) , y_{ss} \Big)\\
        &  y_{ss} = (\epsilon_1 == \epsilon_2) 
\end{aligned}
\end{equation}
\noindent where $L$ is standard binary cross entropy loss and label $y_{ss}$ is $1$ if $\epsilon_1$ is equal to $\epsilon_2$, otherwise $0$. During training with the above self-supervision loss, generator $G$ and the auxiliary network $A$ are updated simultaneously alternating with discriminator updates. Thus, the training objective of LT-GAN is: 
\begin{equation}
    \begin{aligned}
    L_G : &  -\mathbb{E}_{\substack{ z \sim p(z) \\ \epsilon \sim p(\epsilon)} } [D(G(z)) + D(\mathcal{T_{\epsilon}}(G(z))) ] + \lambda . L_A \\
    L_D : & \;\mathbb{E}_{x  \sim p(x) } [1 - D(x) ]_{+} \; \; + \\ & \; \mathbb{E}_{ \substack{ z \sim p(z) \\ \epsilon \sim p(\epsilon) } } ( [1 + D(G(z)) ]_{+} + [1 + D(\mathcal{T_{\epsilon}}(G(z))) ]_{+} ) \\
\end{aligned}\label{eq:lt-loss}
\end{equation}
Here, $\lambda$ denotes the weightage of self-supervision loss in generator. We choose $p(z)$ and $p(\epsilon)$ both to be a normal distribution with standard deviation $\sigma_{z}$ and $\sigma_{\epsilon}$ respectively, where $\sigma_{\epsilon} < \sigma_{z}$. The function $f$ in Eq. $4$ is implemented as difference of encoded features and $E(G(z))$ features are chosen as the intermediate layer activation of the discriminator. Furthermore, in order to balance the min-max training between the generator and the discriminator, we also train the discriminator to predict fake on GAN-induced transformations. An overview of the generator training in LT-GAN is shown in Fig. \ref{fig:flowchart} and pseudo-code of LT-GAN training is explained in Algorithm \ref{lt_gan_algo}.

% The auxiliary task is introduced after $n$ warmup iterations of training using the standard GAN loss to ensure that the generated images and its transformations are closer to natural image manifold.

\begin{algorithm}[t]
\setstretch{1.15}
\SetAlgoLined
\Begin{
\footnotesize{
    \textbf{Input}: Generator, Discriminator and Auxiliary network parameters $\theta_{G},\theta_{D}$ and $\theta_{A}$. Batch size $2b$, weight of self-supervision loss $\lambda$, standard deviation $\sigma_{\epsilon}$ of normal distribution $p(\epsilon)$, discriminator update steps $d_{step}$ for each generator update, Adam hyperparemters $\alpha, \beta_1 , \beta_2$.
    
    \For{\text{number of training iterations}}
    {
        \For{$t:1 ... d_{step}$}
            {   Sample batch $x \sim p_{data}(x) $ \\
                Sample $ \{z^{(i)} , \epsilon^{(i)} \}_{i=1}^{b} \sim p(z) , p(\epsilon)$ \\
                $z = \{z^{(i)}\}_{i=1}^{b} \cup \{z^{(i)} + \epsilon^{(i)}\}_{i=1}^{b}$ \\
                $L_{D} = [1-D(x)]_{+} + [1+D(G(z))]_{+}$ \\
                Update $\theta_{D} \leftarrow
                Adam( L_{D} , \alpha, \beta_1 , \beta_2) $
            }
        
        Sample $z = \{z^{(i)} \}_{i=1}^{2b} \sim p(z)$ , $\epsilon_1, \epsilon_2 \sim p(\epsilon) $ \\
        $\epsilon = [\epsilon_1 ,\epsilon_2].repeat(b)$ \Comment{\textit{repeat along batch dimension}} \\
        Generate images $G(z)$ \\ 
        Generate GAN-induced transformation $G(z + \epsilon)$  \\
        $f(z, z+\epsilon) = E(G(z)) - E(G(z+\epsilon))$ \\
        $shuffle() = $ permutation($2b$)\\
        $L_{A} = L \big( A([ f(z,z+ \epsilon) , f(z,z+ \epsilon).shuffle() ]) , y_{ss} \big) $ \\
        $y_{ss} = (\epsilon == \epsilon.shuffle())$ \\
        $L_{G} = - D(G(z)) - D(G(z+\epsilon))$ \\
        Update $\theta_{A} \leftarrow Adam( L_A , \alpha, \beta_1 , \beta_2 )$ \\
        Update $\theta_{G} \leftarrow
        Adam( (L_G + \lambda . L_A) , \alpha, \beta_1 , \beta_2 ) $
        
    }

    }
 }
 \caption{\footnotesize{Latent Transformation GAN (LT-GAN)}}
\label{lt_gan_algo}
\end{algorithm}

% \subsection{Consistency Regularization in LT-GAN}

\section{Experiments and Results}
\vspace{-5pt}
\paragraph{Datasets} 
We validate our proposed self-supervised task on CIFAR-10~\cite{krizhevsky2010cifar}, STL-10~\cite{stl10}, CelebA-HQ-128~\cite{pggan2018celeba} and ImageNet-2012~\cite{krizhevsky2012imagenet} datasets. CIFAR-10 consists of 60K 32$\times$32 images, belonging to 10 different classes: 50K images for training and 10K for testing. In STL-10, we use 100K unlabelled images for training (resized to 48 $\times$ 48) and 8K images for testing. CelebA-HQ-128 (CelebA-HQ) consists of 30K 128$\times$128 face images. We randomly sample 3K images for testing and the rest for training. ImageNet-2012 consists of approximately 1.2 million images which we downsample to 128-128 resolution for our experiments. We use the 50K validation set images of ImageNet for testing. 

\paragraph{GAN Architectures and Evaluation}
We use the GAN architecture of BigGAN~\cite{biggan2018brock}, StyleGAN~\cite{stylegan2019karras}, SNDCGAN~\cite{sngan_proj} with their proposed training techniques as the baseline. We also compare against state-of-the-art training technique CR-GAN~\cite{crgan2019chen}. In the conditional setting, we perform experiments on CIFAR-10 and ImageNet-2012 with BigGAN architecture. In the unconditional setting, we perform experiments on CelebA-HQ-128 with StyleGAN and SNDCGAN, on CIFAR-10 with SNDCGAN and STL-10 with ResNet architecture \cite{sngan_proj}. 
\par
We use Fr\'echet Inception Distance (FID)~\cite{fid2017martin} as the primary metric for evaluating image quality and diversity. FID has been shown to be more consistent with human evaluation of image quality and also helps in detecting intra-class mode collapse \cite{fid2017martin}. We calculate FID between test set images and equal number of generated images for all datasets and report the best FID obtained across 3 runs. We found our methodology to be stable and we show the variance analysis of FID in the supplementary section. 

\subsection{Training and Implementation Details}
% aux network architecture, aux_lr, eps_std, lambda_bce_loss, g_bs, d_bs, warmup epoch, D_intermediate_layer, lrs, optimizer beta_1 and beta_2, loss function, perform aux task both on baseline and crgan. Aux all pairs loss function. 
The architecture of the auxiliary network $A$ used for the self supervision task consists of a two-layer fully connected network with ReLU activation at the hidden layer and sigmoid activation at the output. Let the features $E(G(z))$ extracted from the discriminator network be of shape $C\times H\times W$. The input layer of $A$ is of $2\times C\times H\times W$ dimension (since we flatten and concatenate the features) and the hidden layer is of $C$ dimension. The self-supervised task is introduced after $n$ warmup iterations of training using the standard GAN loss to ensure that the generated images and its transformations are closer to natural image manifold. Furthermore, we only experimented with sampling two distinct $\epsilon$ and repeating along the batch dimension for calculating GAN-induced transformation. We leave exploring the effect of varying the above number and relaxing the strict equality between $\epsilon$ while calculating self-supervision loss in eq. 5 for the future work. 
% On introducing the self supervision task after $n$ warmup iterations we also adjust the batch size of the generator to twice of that of the discriminator. This helps in providing a larger batch size to train the generator and the auxiliary network on the self supervision task. 
Across all model architectures and datasets, we observe the optimal value of $\sigma_\epsilon$ to lie in the range of $[0.4, 0.6]$. For hyper-parameter $\lambda$, we found the value of $1.0$ to work well except for BigGAN on ImageNet and StyleGAN on CelebA-HQ where we use the value of $0.5$. More details about training hyper-parameters for each dataset and architecture are mentioned in the supplementary. 
% Parth: I think we should either use self supervision task at all places or auxiliary task. Is it okay to mix them up at different places? 
\par 

We use Adam optimizer in all our experiments and spectral normalization (SN)~\cite{sngan_proj} in the discriminator (except in the case of StyleGAN). Hinge loss is used by default for training (except in case of StyleGAN, which uses non-saturating loss with $R_1$ regularization~\cite{R1_reg}). We follow the default configuration for all architectures and hence we train till 200k generator steps for CIFAR-10 and STL-10, 100k generator steps for CelebA-HQ on SNDCGAN, 525K generator steps on StyleGAN. For ImageNet, we train the model for 250K steps unless the training collapses. 
% By default, a learning rate of $0.001$ is used for the auxiliary network and the weight factor $\lambda$ of the auxiliary loss term in the generator objective function is set to $1.0$. 
% \par
% Adam optimizer is used for all our experiments. By default, spectral normalization (SN)~\cite{} is used in the discriminator (except in the case of StyleGAN). We stop training after 200k generator steps for CIFAR-10, 100k generator steps for CelebA-HQ-128 on SNDCGAN, and 525K generator steps for CelebA-HQ-128 on StyleGAN. By default, hinge loss is used for the generator and the discriminator (except in case of StyleGAN, which uses non-saturating loss with $R_1$ regularization).
% By default, a learning rate of $0.0002$ is used for the generator and the discriminator (except for StyleGAN, where we use $0.0015$). By default, hinge loss is used for the generator and the discriminator (except in case of StyleGAN, which uses non-saturating loss with $R_1$ regularization).

\par
In the following sections, we show that our proposed self-supervision task helps in improving FID scores over the baseline models and can be effectively combined with other regularization techniques in GANs, e.g. CR-GAN \cite{crgan2019chen}, across datasets and model architectures. We empirically show that LT-GAN results in a more steerable and disentangled latent space by performing latent space manipulation on CelebA-HQ and ImageNet datasets. We also compare our approach with the recently proposed self-supervised SS-GAN, which uses a rotation-based auxiliary task \cite{rot_ssgan2019chen}. 

\begin{table}[t]
\begin{tabular}{lll}
\hline
DATASET   & METHOD               & FID    \\ \hline
          & BigGAN       & 14.73  \\
CIFAR-10  & LT-BigGAN (ours)    & 11.01    \\
(cond.)   & CR-BigGAN     & 11.48   \\
          & CR+LT-BigGAN (ours) & \textbf{9.80}   \\ \hline
          & SNDCGAN        & 25.39  \\
CIFAR-10  & LT-SNDCGAN (ours)    & 22.10   \\
(uncond.) & CR-SNDCGAN   & 18.72  \\
          & CR+LT-SNDCGAN (ours) & \textbf{17.56}   \\ \hline
          & SNDCGAN              & 25.95  \\
          & LT-SNDCGAN  (ours)         & 19.63  \\
          & CR-SNDCGAN          & 16.97 (18.44$^{*}$) \\
CelebA-HQ & CR+LT-SNDCGAN (ours)      & \textbf{16.84}  \\ \cline{2-3}
(uncond.)          & StyleGAN             & 11.43  \\
          & LT-StyleGAN (ours)         & 11.15  \\ \hline
ImageNet  & BigGAN        & 10.34$^{^{\#}}$ \\
(cond.)   & LT-BigGAN (ours)    & 9.94$^{^{\#}}$  \\ \hline
\end{tabular}
\caption{\footnotesize{Comparison of self-supervised LT-GAN training approach with state-of-the-art GANs based on FID. * denotes our best reproduced result using the implementation of $^{1}$, which is different from the score reported in \cite{crgan2019chen}. $^{\#}$ denotes BigGAN Imagenet implementation of $^{2}$}}
\label{tab:sota-table}
\end{table}

\begin{figure*}[t]
\centering
    \begin{minipage}{.45\textwidth}
      \includegraphics[width=\linewidth]{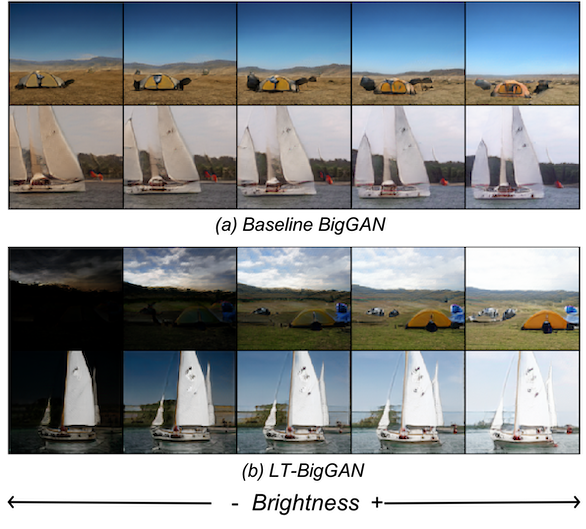}
    \end{minipage}%
    \hspace{0.5cm}
    \begin{minipage}{.45\textwidth}
      \includegraphics[width=\linewidth]{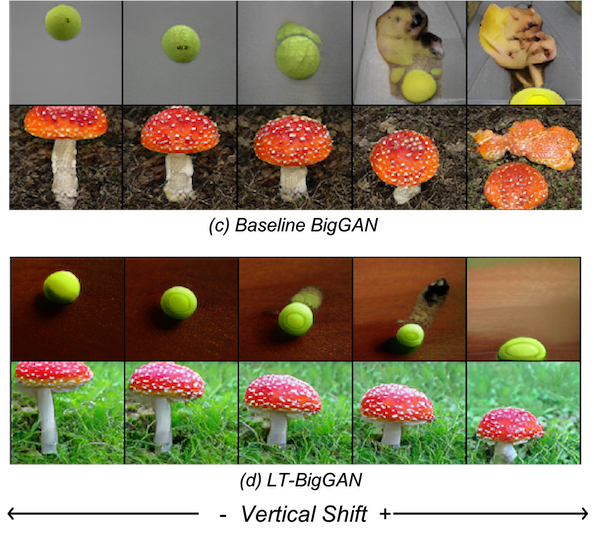}
    \end{minipage}
    \caption{\footnotesize{Qualitative comparison of Brightness (left) and Vertical shift (right) using latent space manipulation on randomly generated images of ImageNet for Baseline BigGAN and LT-BigGAN model.}}
    \label{fig:brightness}
\end{figure*}

\subsection{Results}

\paragraph{Unconditional GANs}
In the unconditional setting, we perform experiments on CIFAR-10 with SNDCGAN \cite{sngan_proj} architecture and CelebA-HQ with SNDCGAN and StyleGAN \cite{stylegan2019karras} architectures \footnote{We use the open source implementation of \url{https://github.com/google/compare\_gan} for SNDCGAN and \url{https://github.com/rosinality/style-based-gan-pytorch} for StyleGAN}. 

From Table \ref{tab:sota-table}, we can see that LT-GAN results in improved FID scores compared to the baseline models. Moreover, we also combine our self-supervision task with the current state of the art training methodology CR-GAN \cite{crgan2019chen}. Using CR+LT-GAN results in further improvement of FID score on both datasets.

\paragraph{Conditional GANs}
In the conditional setting, both the generator and the discriminator are provided with the underlying class labels information. We perform experiments on CIFAR-10 and ImageNet datasets \footnote{Experiments on BigGAN use the implementation of  \url{https://github.com/ajbrock/BigGAN-PyTorch}. For ImageNet $d_{step}$ is $1$ instead of the more optimal setting of $2$ \cite{biggan2018brock} because of less computation requirements of the former.} using the recent state-of-the-art BigGAN \cite{biggan2018brock} model. We observe that for both datasets our self-supervision task improves the FID score as shown in Table \ref{tab:sota-table}. We also present experimental results of combining our self-supervision technique with the current state-of-the-art CR-GAN \cite{crgan2019chen} on CIFAR-10 that further improves the FID score over CR-GAN.

\subsection{Steerability in Latent Space}
In this section, we empirically demonstrate that our proposed self-supervision task helps to learn a more steerable latent space. We analyse models trained on CelebA-HQ dataset using the framework of InterfaceGAN~\cite{interface2020shen}. On ImageNet dataset, we use the methodology proposed in~\cite{controlling2020iclr} to show that BigGAN model \cite{biggan2018brock} trained with our approach LT-GAN helps in finding better edit directions in the latent space corresponding to image transformations like translation, brightness and scale.

% \vspace{-12pt}
\paragraph{ImageNet Dataset}
We analyze the latent space structure of generator trained on ImageNet by finding interpretable directions corresponding to parametrizable continuous factors of variation like translation, zoom and brightness using the framework of \cite{controlling2020iclr}. To this end, authors in ~\cite{controlling2020iclr} propose a novel reconstruction loss between randomly generated images and the transformed version of these images with varying intensity level (e.g. zoom at various scales) to first determine the latent code corresponding to transformed images. Using this training set of pair of latent codes of original and transformed images, the direction in latent space corresponding to that particular transformation is learned as explained in \cite{controlling2020iclr}. We use this methodology to discover latent space trajectories corresponding to the image transformations: brightness, scale, horizontal shift and vertical shift for BigGAN \cite{biggan2018brock} conditional model trained on ImageNet dataset.

% One  of  the  most  intuitive  ways  to  identify  latent  spacetrajectories  corresponding  to  image  transformations  liketranslation, zoom, brightness, etc. is to somehow determine the  latent  code  corresponding  to  the  transformed  image.We  can  then,  compute  its  difference  with  the  latent  codeof  the  original  image  to  find  the  latent  space  directioncorresponding to the image transformation.  To this end,  []proposes  a  novel  reconstruction  loss  to  invert  generativemodels.    The  new  loss  function  reduces  the  importancegiven to higher frequencies while inverting the generativemodel  and  hence,  results  in  sharper  and  more  realisticimages.  We use this loss function to find the latent codescorresponding  to  the  transformed  images  and  hence  dis-cover latent space trajectories corresponding to brightnessand scale for BigGAN models trained on ImageNet dataset.

Fig.\ref{fig:brightness} shows the qualitative comparison of brightness and vertical shift direction vectors between baseline BigGAN and LT-BigGAN. We can see in the figure that our approach results in smoother and more meaningful transformations in the image space while preserving the content of the image and avoiding distortions at the extremes. More qualitative comparison on latent space steerability including horizontal shift and zoom is shown in the supplementary.

\begin{figure*}[t]
\centering
    \begin{minipage}{.45\textwidth}
      \includegraphics[width=\linewidth]{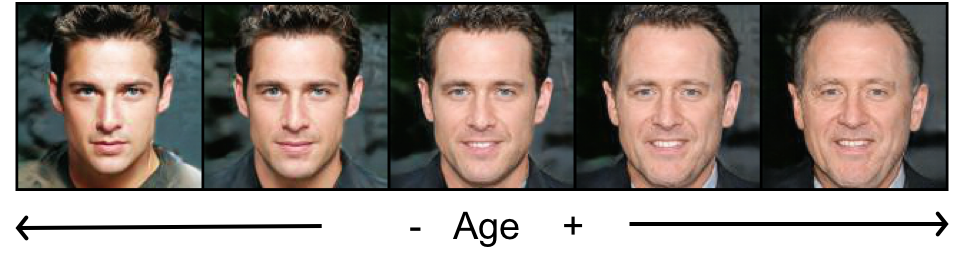}

    %   \subcaption{}
    \end{minipage}%
    \hspace{0.5cm}
    \begin{minipage}{.45\textwidth}
      \includegraphics[width=\linewidth]{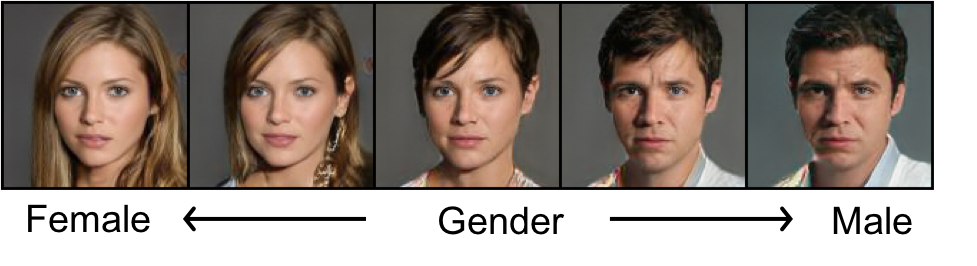}

    %   \subcaption{}
    \end{minipage}\\
    \vspace{0.2cm}
    \begin{minipage}{.45\textwidth}
      \includegraphics[width=\linewidth]{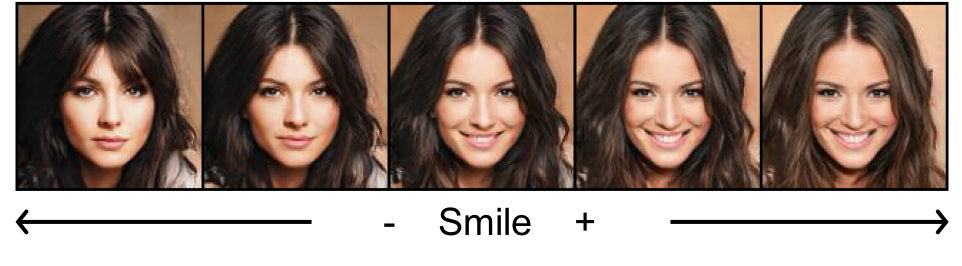}

    %   \subcaption{}
    \end{minipage}%
    \hspace{0.5cm}
    \begin{minipage}{.45\textwidth}
      \includegraphics[width=\linewidth]{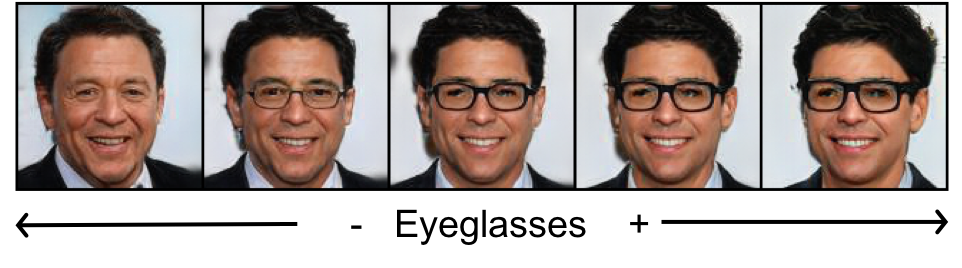}

    %   \subcaption{}
    \end{minipage}\\
    \caption{\footnotesize{Manipulation of Age(top left), Gender(top right), Smile(bottom left)and Eyeglasses(bottom right) attributes by navigating the latent space of LT-StyleGAN using InterfaceGAN \cite{interface2020shen} framework. Original images are in the centre and the left and right images are generated by moving the latent code in negative and positive directions respectively.}}
    \label{fig:lt-stylegan}
\end{figure*}

\vspace{-8pt}
\paragraph{CelebA-HQ Dataset}\label{sec:steer}
InterfaceGAN~\cite{interface2020shen} provides 
a framework to find the interpretable semantic directions encoded in the latent space of face synthesis GAN models. Using it, we discover directions in the latent space to smoothly vary facial attributes, namely age, gender, smile expression and eyeglasses. We use the following procedure as proposed in \cite{interface2020shen} to discover facial attribute boundaries for StyleGAN and SNDCGAN architectures:
\begin{itemize}
  \item Randomly generate 500K images and use a ResNet50 facial attribute detector to predict the value of each binary facial attribute for the generated images. For each binary attribute, sort the list of 500k images based on the predicted value of the attribute and collect top 10k and bottom 10k images. Out of these 20k images, randomly sample 14k images to use as the training set.
  \item For each attribute, train a linear SVM using the above collected 14k images to identify the value of the attribute (i.e. 0/1) given the latent code that was used to generate the image. The trained SVM represents a hyperplane that serves as a boundary in the latent space separating the two (-ve/+ve) classes of the binary attribute.
\end{itemize}
\par
We report the accuracy of each of the trained SVMs on remaining set of images (i.e. 480k images). A higher SVM accuracy indicates a more steerable latent space. As shown in Table \ref{tab:interfacegan-table}, our self-supervision task improves upon the baseline accuracy on all four facial attributes (i.e. age, eyeglasses, gender and expression) for both SNDCGAN and StyleGAN architectures. We also compare LT-GAN against SSGAN, which is another self-supervision based GAN. LT-GAN achieves better accuracy on all four attributes compared to SSGAN~\cite{rot_ssgan2019chen}. Fig.\ref{fig:lt-stylegan} shows some example images of attribute manipulation by moving the latent code in the direction normal to attribute boundary. We show more qualitative samples and its comparison with the baseline model in the supplementary section.

% \begin{table}[t]
% \centering
% \begin{tabular}{|l|lll|}
% \hline
% \multirow{2}{*}{Attributes} & \multicolumn{3}{c|}{SNDCGAN} \\
%                             & Baseline  & SS-GAN  & LT-GAN      \\ \hline
% Age                         & 63.89     & 67.28   & \textbf{71.01}      \\
% Eyeglasses                  & 77.36     & 82.85   & \textbf{88.53}        \\
% Gender                      & 64.64     & 68.34   & \textbf{72.06}      \\
% Smile                       & 86.10     & 85.76   & \textbf{88.55}    \\ \hline
% & \multicolumn{3}{c|}{StyleGAN} \\
% & Baseline       & LT-GAN   & \\ \hline
% Age & 68.66          & \textbf{70.57}  &  \\
% Eyeglasses & 70.95          & \textbf{76.91 }&  \\
% Gender & 73.78          & \textbf{78.49}  &  \\
% Smile & 64.75          & \textbf{65.30}  &   \\ \hline
% \end{tabular}
% \caption{SVM accuracy (\%) on remaining set (i.e. 480K images) for facial attributes age, eyeglasses, gender and smile.}
% \label{tab:interfacegan-table}
% \end{table}

\begin{table}[t]
\centering
\resizebox{\columnwidth}{!}{%
\begin{tabular}{|l|lll|ll|}
\hline
\multirow{2}{*}{} & \multicolumn{3}{c|}{SNDCGAN} & \multicolumn{2}{c|}{StyleGAN} \\
                            & Baseline  & SS-GAN & LT-GAN  & Baseline  & LT-GAN      \\ \hline
A    & 63.89 & 67.28  & \textbf{71.01}  & 68.66  & \textbf{70.57}    \\
E   & 77.36 & 82.85  & \textbf{88.53}   & 70.95 & \textbf{76.91}     \\
G     & 64.64 & 68.34  & \textbf{72.06}   & 73.78  & \textbf{78.49}   \\
S      & 86.10 & 85.76  & \textbf{88.55}  & 64.75  & \textbf{65.30}  \\ \hline
\end{tabular}
}
\caption{\footnotesize{Classification accuracy (\%) on separation boundaries in
latent space with respect to different attributes of CelebA-HQ. Attributes are A: Age, E: Eyeglasses, G: Gender, and S: Smiling.}}
\label{tab:interfacegan-table}
\end{table}

% \begin{figure}[t]
% \centering
% % \fbox{\rule{0pt}{2in} \rule{0.9\linewidth}{0pt}}
%   \includegraphics[width=0.8\linewidth]{images/brightness_ours.png}
%   \caption{Change in brightness using common $w$ learned on a set of training images}
% \label{fig:long}
% \label{fig:onecol}
% \end{figure}

% \begin{table}[]
% \begin{tabular}{|l|l|l|l|}
% \hline
%                           &          & SNDCGAN & ResNet \\ \hline
% \multirow{3}{*}{CIFAR-10}  & Baseline &    25.39     &   16.20     \\  
%                           & SS-GAN   &  23.39       &    14.60    \\ 
%                           & LT-GAN   &   22.10      &    15.30   \\  \hline
% \multirow{3}{*}{STL-10}  & Baseline &    40.30     &   35.74     \\ 
%                           & SS-GAN   &  36.30      &    33.63    \\ 
%                           & LT-GAN   &   36.18      &    31.35   \\  \hline
% \end{tabular}
% \caption{Comparison of LT-GAN with SS-GAN}
% \label{tab:my-table}
% \end{table}

\section{Discussion and Ablation Studies}
% \subsection{Disentangled attribute vectors in models using correlation analysis}
\paragraph{StyleGAN Latent Space Disentanglement}
\par
To demonstrate that our proposed self-supervision task helps in achieving a more disentangled latent space, we adopt the InterFaceGAN framework ~\cite{interface2020shen} to measure the correlation between synthesized facial attributes distributions of StyleGAN trained on CelebA-HQ dataset. We synthesize 500K images by randomly sampling the latent space. Using a pre-trained ResNet50 facial attribute detector, we assign attribute scores to all 500K images for all four facial attributes (age, eyeglasses, gender, and smile). Treating each attribute score as a random variable, we can compute the correlation between two attributes using their distribution observed over the 500K generated images. The formula to compute correlation between two attributes $X$ and $Y$ is $\rho_{XY} = \dfrac{Cov(X, Y)}{\sigma_X\sigma_Y}$, where $Cov(\cdot,\cdot)$ denotes covariance and $\sigma$ denotes standard deviation. Correlation values closer to zero indicate a more disentangled latent space. Table \ref{tab:correlation-syn-attr-dis-stylegan} shows the correlation values between attributes for both baseline StyleGAN and LT-StyleGAN. It can be observed that the correlation between attributes is more closer to $0$ for LT-StyleGAN as compared to baseline StyleGAN. 

Similar to \cite{stylegan2019karras}, we also compute the perceptual path length for both latent spaces $Z$ and $W$ of StyleGAN. The idea is that a more disentangled latent space will result in perceptually smoother transitions in the image space as we interpolate in the latent space, and thus give lower perceptual path length. For the $Z$ space, perceptual path length is 242.33 for baseline StyleGAN and 133.11 for LT-StyleGAN. For the $W$ space, perceptual path length is 77.48 for baseline StyleGAN and 72.71 for LT-StyleGAN.

\begin{figure}[t]
\centering
    \includegraphics[width=\linewidth]{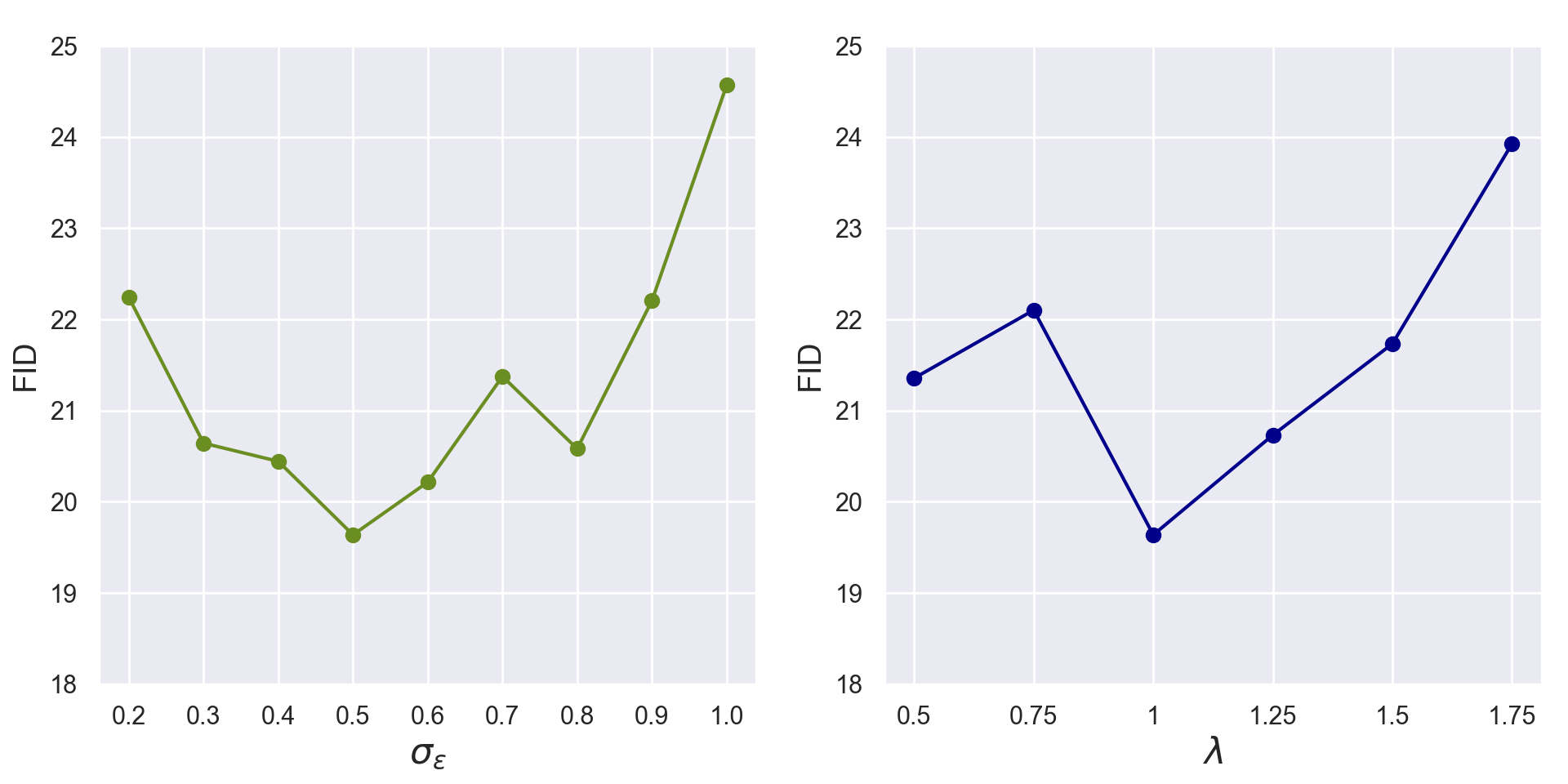}
    \caption{\footnotesize{FID on varying $\sigma_{\epsilon}$ and $\lambda$ for LT-SNDCGAN on CelebA-HQ}}
    \label{fig:ablation_hp}
\end{figure}

\begin{table}[t]
\begin{tabular}{|c|c|c|c|c|}
\hline
           & A     & E  & G      & S       \\ \hline
A        & 1./1. & 0.373/\textbf{0.326} & 0.466/\textbf{0.462} & -0.128/\textbf{-0.111} \\ \hline
E & -       & 1./1.     & 0.292/\textbf{0.262} & -0.096/\textbf{-0.088} \\ \hline
G     & -       & -           & 1./1.     & -0.297/\textbf{-0.293} \\ \hline
S    & -       & -           & -           & 1./1.       \\ \hline
\end{tabular}
\caption{\footnotesize{Correlation matrix of synthesized attribute distributions of StyleGAN on CelebA-HQ. In each cell, the first value corresponds to baseline StyleGAN and the second value (following /) corresponds to LT-StyleGAN. Attributes are A: Age, E: Eyeglasses, G: Gender, and S: Smiling.}}
\label{tab:correlation-syn-attr-dis-stylegan}
\end{table}
\vspace{-8pt}
\paragraph{Choice of hyper-parameter $\sigma_{\epsilon}$ and $\lambda$.} 
Hyper-parameter $\sigma_{\epsilon}$ controls the difficulty of self-supervision task. A large value of $\sigma_\epsilon$ makes the self-supervision task trivial (since it is easier to distinguish between latent space perturbations that are far apart). In contrast, a smaller value of $\sigma_\epsilon$ makes the pretext task too difficult and may cripple training. Hyper-parameter $\lambda$ controls the ratio of weight assigned to self-supervision loss and adversarial loss in generator objective function. To study the effect of these hyper-parameters on model performance (i.e. FID), we perform ablation experiments by varying one hyper-parameter and fixing the other to its optimal value. We conduct this experiment on SNDCGAN architecture with CelebA-HQ dataset and the results are as shown in Fig. \ref{fig:ablation_hp}. It can be observed that minimum FID is achieved at the optimal values of $\sigma_\epsilon=0.5$ and $\lambda=1.0$. FID increases as we move away from the optimal values and the graphs show a U-shaped trend.

% Thus, it is important to correctly balance the difficulty of the self supervision task and choose an optimal value of $\sigma_\epsilon$
% we balance it such that it neither outweighs nor is outweighed by the standard GAN loss function.
%explain the results using table. 

% \begin{table}[]
% \begin{tabular}{|l|c|c|c|c|c|c|c|l|c|}
% \hline
%                         \sigma_\epsilon & 0.2 & 0.3 & 0.4 & 0.5         & 0.6 & 0.7 & 0.8 & 0.9 & 1.0 \\ \hline
% FID & 22.24  & 20.64  & 20.44  & \textbf{19.63} & 20.22  & 21.37  & 20.58  & \multicolumn{1}{c|}{22.20} & 24.57  \\ \hline
% \end{tabular}
% \end{table}

% \begin{table}[]
% \begin{tabular}{|l|c|l|c|l|c|l|}
% \hline
% \lambda & 0.5 & 0.75                    & 1.0 & 1.25                    & 1.5 & 1.75                    \\ \hline
% FID & 20.50  & 22.18 & \textbf{19.63}  & 20.73 & 23.81  & 23.92 \\ \hline
% \end{tabular}
% \end{table}

% \begin{table}[]
% \begin{tabular}{lccccccccc}
%                          & σ = 0.2 & σ = 0.3 & σ = 0.4 & σ = 0.5         & σ = 0.6         & σ = 0.7 & σ = 0.8 & \multicolumn{1}{l}{σ = 0.9} & σ = 1.0 \\
% SNDCGAN CelebA-HQ LT-GAN & 22.241  & 20.642  & 20.443  & \textbf{19.637} & 20.215          & 21.374  & -       & -                           & 24.567  \\
% SNDCGAN CIFAR-10 LT-GAN  & -       & -       & 23.379  & 22.118          & \textbf{22.109} & -       & 22.706  & 23.890                      & 23.546 
% \end{tabular}
% \end{table}

% \paragraph{Choice of $f$ and architecture of auxiliary network}
% To capture the change in generated image and its transformation, we choose $f$ (described in Eq. 4) as a subtraction of corresponding encoded features. To explore other choice of $f$, we experimented with the concatenation of the features of original and transformed images and found similar performance. For the architecture of auxiliary network, we experimented with 1) Linear network 2) Non-linear network with a hidden layer and ReLU activation. We found that a linear network was unable to differentiate between generative transformations resulting from different $\epsilon$'s and the accuracy of the auxiliary task remained close to $50\%$ throughout training. Non linear network with single hidden layer worked well across all datasets and architectures with prediction accuracy of $>90\%$ by the end of training. 
% % We believe this is due to the different capacities of both the networks and the introduction of non-linearity in the two layer auxiliary network. 
\vspace{-8pt}

\begin{figure}[t]
\centering
    \includegraphics[width=\linewidth, height=4.1cm]{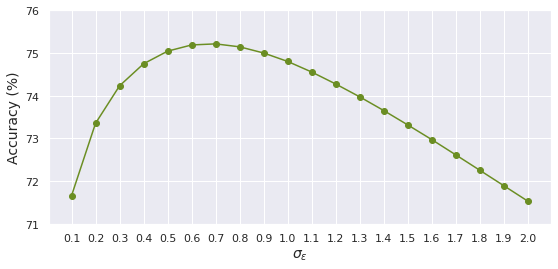}
    \caption{\footnotesize{Accuracy (\%) of our binary self-supervision task on varying $\sigma_\epsilon$ for LT-SNDCGAN on CelebA-HQ}}
    \label{fig:eps_acc_plot}
\end{figure}

\paragraph{Auxiliary network accuracy on generative transformations}
We validate the efficacy of learned auxiliary network $A$ in SNDCGAN CelebA-HQ setting with $\sigma_{\epsilon}$ = $0.5$. We vary the $\sigma_{\epsilon}$ of $p(\epsilon)$ and test the ability of the auxiliary network to distinguish between these GAN-induced transformations. In Fig. \ref{fig:eps_acc_plot}, we report the binary classification accuracy on random 25K generated samples and its transformation from the trained generator. It can be observed that the auxiliary model classifies relatively well for transformations with $\sigma_{\epsilon}$ in the neighbourhood of $0.5$ on which it was trained, but the performance decreases as $\sigma_{\epsilon}$ diverges from $0.5$.
% as $\sigma_{\epsilon}$ gets closer to 0. 

% Use of Aux network on test dataset to show that it actually generalizes. Also generate attention maps highlighting the changes in generated images by same of different perturbations is made using aux sub-network. 

% \begin{table}[]
% \begin{tabular}{|l|l|l|l|}
% \hline
%         &    CELEBA-HQ  & CIFAR-10  &  STL-10 \\ \hline
% Baseline &    25.95     & 25.39     &   40.30    \\  
% SS-GAN   &   26.85      & 22.88     &   36.30    \\ 
% LT-GAN   &   \textbf{19.63}      &  \textbf{22.10}   &    \textbf{36.18}   \\ \hline

% \end{tabular}
% \caption{Comparison of LT-GAN with SS-GAN on SNDCGAN architecture}
% \label{tab:ssgan}
% \end{table}

\begin{table}[t]
\centering
\begin{tabular}{|l|l|l|l|}
\hline
 Methods  &    CelebA-HQ  & CIFAR-10  &  STL-10 \\ \hline
Baseline &    25.95     & 25.39     &   35.74    \\  
SS-GAN   &   26.85      & 22.88     &   33.63    \\ 
LT-GAN (ours)   &   \textbf{19.63}      &  \textbf{22.10}   &    \textbf{31.35}   \\ \hline

\end{tabular}
\caption{\footnotesize{FID comparison of LT-GAN with SS-GAN on different datasets}}
\label{tab:ssgan}
\end{table}

\vspace{-14pt}
\paragraph{Comparison with SS-GAN}
We also compare LT-GAN with SS-GAN \cite{rot_ssgan2019chen} which is a recently proposed technique to train GAN's with rotation self-supervision. In contrast to SS-GAN our self-supervision task is only defined with respect to generator and considers generative transformations instead of rotation transformation. We compare across 3 datasets: CIFAR-10 and CelebA-HQ on SNDCGAN architecture and STL-10 on resent architecture \cite{sngan_proj}. The results are shown in Table \ref{tab:ssgan}. We observe that LT-GAN performs better on CELBA-HQ and STL-10 and is comparable to SS-GAN on CIFAR-10. Since rotation transformation is less informative for datasets with single domain images like faces, SS-GAN performs worse than baseline on CelebA-HQ dataset. However in comparison to SS-GAN, LT-GAN improves the FID score for all datasets. 

\paragraph{Classification Accuracy Score (CAS)}
CAS \cite{cas2019metric} was recently proposed as an additional metric for evaluating conditional generative models on the downstream task of image classification. A standard image classification network is trained using images generated from the model as a training set. The trained model is used to predict labels on the testing set of real images and the obtained test accuracy is the CAS metric (higher the better). It was shown that neither FID \cite{fid2017martin} nor IS\cite{inception2016} scores are predictive of CAS, and thus it serves as another independent evaluation metric. We compare the CAS score of LT-BigGAN and baseline BigGAN model on Imagenet and CIFAR-10 dataset. For ImageNet dataset, we trained a ResNet-50 \cite{resnet2015He} classifier similar to \cite{cas2019metric} using the generated samples and evaluated its performance on its validation set. LT-GAN achieves the top-5 accuracy of $49.24$ as compared to $44.15$ of the baseline model, outperforming it by over $5\%$. Furthermore, on closer examination of class level classification accuracy, we found that many classes in the baseline model suffer from severe mode collapse, which is alleviated to a large extent in LT-GAN. Sample images from those classes for both baseline and LT-GAN are shown in supplementary section Fig. \ref{fig:lt-biggan_imagenet_mode}. On CIFAR-10 dataset, a ResNet-56 model (as used in \cite{cas2019metric}) trained via samples generated from LT-BigGAN achieved the test accuracy of $79.93 \%$ whereas the baseline model achieved $70.57 \%$. 

%%supplementary
%hyperparameter for each model and dataset
%validation set images of steerability on imagenet %mode collapse images
%CELEBA-hq images 
%what features are captured in the hidden layer of aux network. backpropagation from hidden layer?
\section{Conclusion}
In this work, we present LT-GAN, a novel self-supervised technique for improving the diversity and quality of GAN’s image generation. The pretext task of identifying GAN-induced transformation helps the generator blocks of GANs to learn steerable latent feature representation and synthesise high-fidelity images. The experimental results demonstrate that when combined along with strong GAN baselines \cite{biggan2018brock, stylegan2019karras}, our model LT-GAN improves the quality and diversity of generated image on several standard datasets. The performance on FID metric and controlled image editing highlights the effectiveness of LT-GAN in unconditional and class conditional GAN settings. We hope that this approach of leveraging latent transformation as a pretext task can be extended to other generative models.

{\small
\bibliographystyle{ieee_fullname}
\bibliography{egbib}
}

\twocolumn[\appendixhead]

% In Section \ref{sec:more_results}, we present some additional results on Inception Score \cite{inception2016}, variance in FID \cite{fid2017martin} across multiple runs and ablation study on architecture of auxiliary network. In Section \ref{sec:mode_collapse}, we show qualitative comparison between baseline BigGAN and LT-BigGAN ImageNet model on latent space manipulation and classes with mode collapse. Section \ref{sec:stylegan} shows more examples images of facial attribute editing in LT-StyleGAN. Lastly, section \ref{sec:hyperparameter} provides the hyper-parameter details used in training LT-GAN over different datasets and architectures. 

\section{Additional Results and Ablation Study}\label{sec:more_results}
% In this section, we present experiments to study the choice of auxiliary network architecture and additional quantitative evaluation of LT-GAN against the baseline GANs. 
% We also conduct some experiments to qualitatively evaluate the diversity and image quality of the generated images.

\begin{figure}[b]
\centering
    \includegraphics[width=\linewidth]{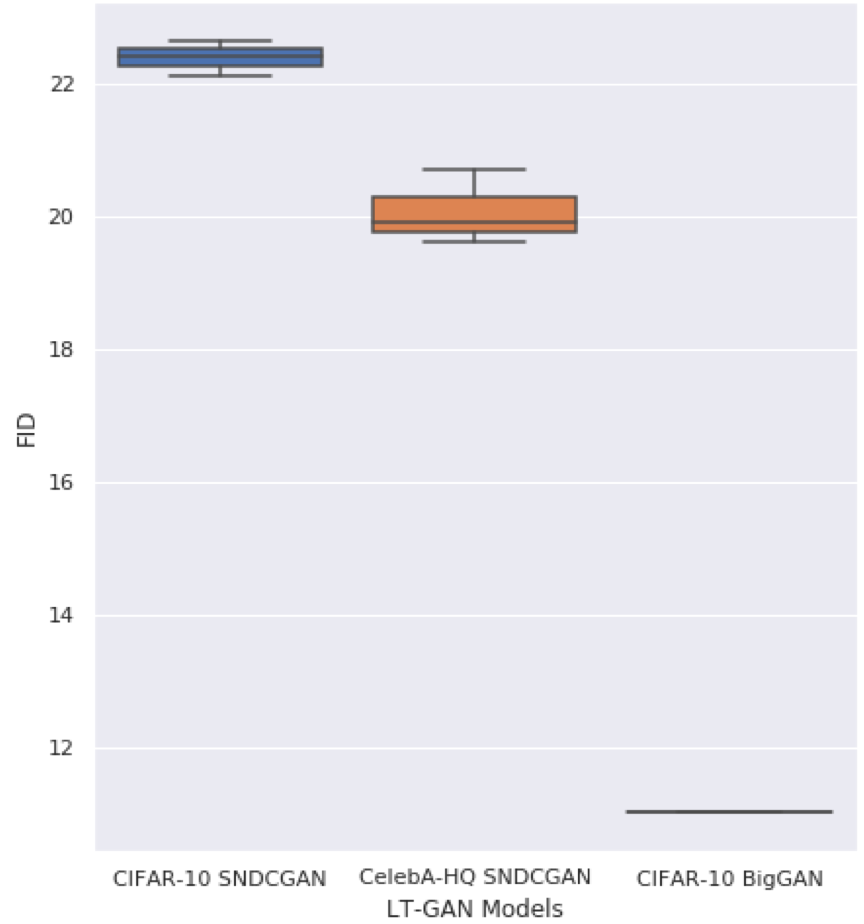} 
    % \hspace{0.1cm}
    \caption{\footnotesize{FID Variance box plot of LT-GAN approach on different architectures for CIFAR-10 and CelebA-HQ.}}
    \label{fig:boxplot}
\end{figure}

\paragraph{Variance analysis of FID}
We show the FID \cite{fid2017martin} variance box-plot of our approach LT-GAN on SNDCGAN \cite{sngan_proj} and BigGAN \cite{biggan2018brock} architectures for CIFAR-10 and CelebA-HQ datasets in Fig \ref{fig:boxplot}. To provide a fair study of FID evaluation on our approach, for each configuration, we compute the FID 3 times with different random initial seeds.
% BigGAN cifar-10: 11.016087, 11.034117, 11.018377
% SNDCGAN CelebA-HQ: 19.634, 19.903, 20.703.
% SNDCGAN CIFAR-10: 22.109, 22.410, 22.657.

\paragraph{Inception Score}
We evaluate our approach of LT-GAN using another GAN evaluation metric named Inception Score (IS) \cite{inception2016}. Here, we report the Inception Score of models trained on CIFAR-10 dataset. As shown in Table \ref{tab:inception_score},  LT-GAN improves IS over baseline, while CR+LT approach achieves the best IS results, on both SNDCGAN and BigGAN architectures.

% SNDCGAN CIFAR-10: I have reported the best results across all experiments conducted: \\
% Baseline: 7.54 \\
% LT-GAN: 7.85 \\
% CR-GAN: 7.93 \\
% CR+LT-GAN: 8.16 \\

% Cifar-10 BigGAN
% Baseline: 8.787805 +/- 0.126509.
% LT-BigGAN: 9.125070 +/- 0.256754.
% CR-GAN: 
% LT+CR-BigGAN: 9.174415 +/- 0.181512.

\begin{table}[b]
\centering
\begin{tabular}{|l|c|c|}
\hline
         & SNDCGAN & BigGAN \\ \hline
Baseline  & 7.54             & 8.79            \\
% \hline
LT-GAN    & \textbf{7.85}             & \textbf{9.13}            \\ \hline
CR-GAN    & 7.93             &  9.17              \\
% \hline
CR+LT-GAN & \textbf{8.16}             & \textbf{9.17}            \\ \hline
\end{tabular}
\caption{\footnotesize{Inception Score for SNDCGAN and BigGAN architectures trained using different approaches on CIFAR-10.}}
\label{tab:inception_score}
\end{table}

\begin{figure*}[t]
\centering
    \begin{minipage}{.32\textwidth}
    \includegraphics[width=\linewidth]{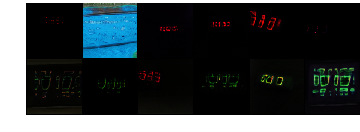}
    \end{minipage}%
    \hspace{0.01cm}
    \begin{minipage}{.32\textwidth}
    \includegraphics[width=\linewidth]{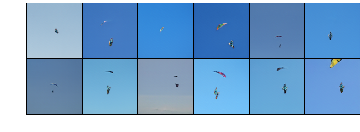}
    \end{minipage}%
    \hspace{0.01cm}
    \begin{minipage}{.32\textwidth}
    \includegraphics[width=\linewidth]{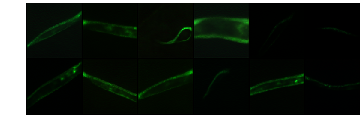}
    \end{minipage} 
    \begin{minipage}{.32\textwidth}
    \includegraphics[width=\linewidth]{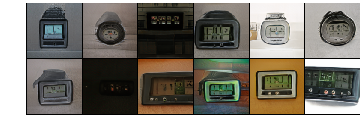}
    \subcaption{}
    \end{minipage}%
    \hspace{0.01cm}
    \begin{minipage}{.32\textwidth}
    \includegraphics[width=\linewidth]{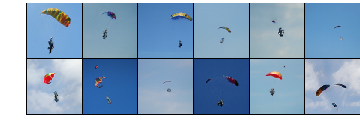}
    \subcaption{}
    \end{minipage}%
    \hspace{0.01cm}
    \begin{minipage}{.32\textwidth}
    \includegraphics[width=\linewidth]{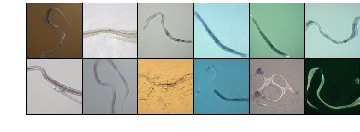}
    \subcaption{}
    \end{minipage} \\

    \begin{minipage}{.32\textwidth}
    \includegraphics[width=\linewidth]{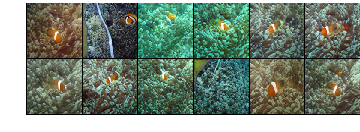}
    \end{minipage}%
    \hspace{0.01cm}
    \begin{minipage}{.32\textwidth}
    \includegraphics[width=\linewidth]{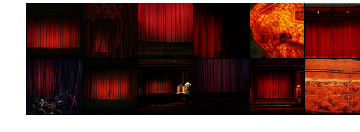}
    \end{minipage}%
    \hspace{0.01cm}
    \begin{minipage}{.32\textwidth}
    \includegraphics[width=\linewidth]{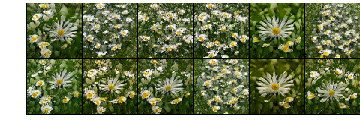}
    \end{minipage} \\
    \vspace{-8pt}
    \begin{minipage}{.32\textwidth}
    \includegraphics[width=\linewidth]{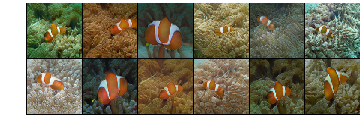}
    \subcaption{}
    \end{minipage}%
    \hspace{0.01cm}
    \begin{minipage}{.32\textwidth}
    \includegraphics[width=\linewidth]{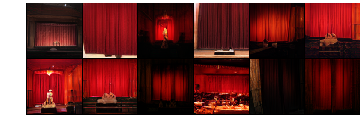}
    \subcaption{}
    \end{minipage}%
    \hspace{0.01cm}
    \begin{minipage}{.32\textwidth}
    \includegraphics[width=\linewidth]{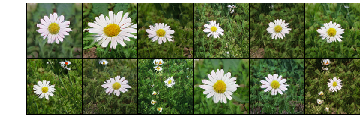}
    \subcaption{}
    \end{minipage} 
    \caption{\footnotesize{Samples of generated images from categories with mode collapse in Baseline BigGAN and its corresponding images generated from LT-BigGAN model. The 6 blocks of images corresponds to ImageNet classes: (a) \textit{digital clock}, (b) \textit{parachute}, (c) \textit{nematode}, (d) \textit{anemone fish} ,(e) \textit{theater curtain}, (f) \textit{daisy}. In each block (that comprises of 4 rows of images), the top part (1st and 2nd row) corresponds to images generated using Baseline (BigGAN) model and the bottom part (3rd and 4th row corresponds to images produced using our approach LT-BigGAN. }}
    \label{fig:lt-biggan_imagenet_mode}
\end{figure*}

\begin{figure*}[t]
\centering
    \begin{minipage}{.48\textwidth}
    \includegraphics[width=\linewidth]{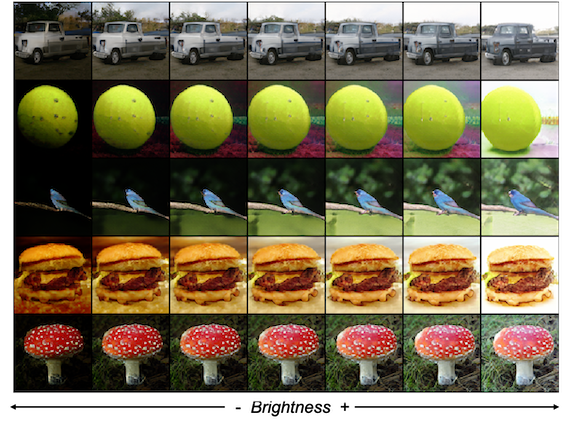}
    \end{minipage}%
    \hspace{0.05cm}
    \begin{minipage}{.48\textwidth}
    \includegraphics[width=\linewidth]{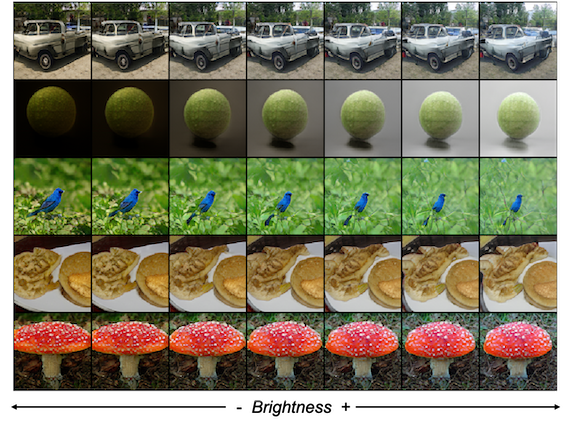}
    \end{minipage}\\
    \begin{minipage}{.48\textwidth}
    \includegraphics[width=\linewidth]{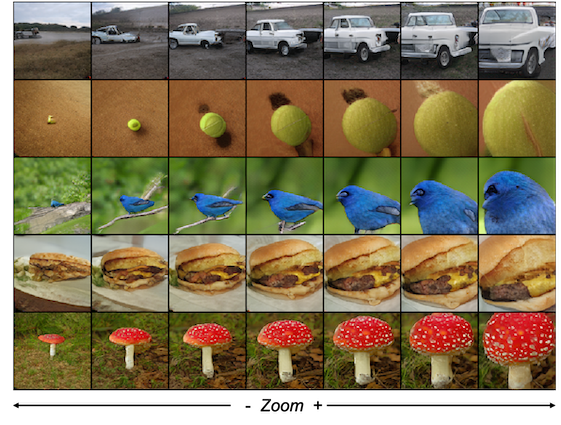}
    \subcaption{LT-BigGAN}
    \end{minipage}%
    \hspace{0.05cm}
    \begin{minipage}{.48\textwidth}
    \includegraphics[width=\linewidth]{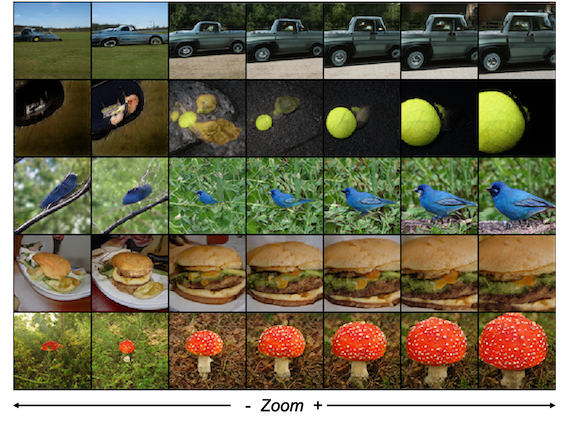}
    \subcaption{BigGAN}
    \end{minipage}
    \caption{\footnotesize{Qualitative comparison for varying brightness and zoom between LT-BigGAN (left) and Baseline BigGAN (right) in five categories of ImageNet through latent space manipulation method of \cite{controlling2020iclr}}}
    \label{fig:lt-biggan_imagenet_scale_zoom}
\end{figure*}

\paragraph{Choice of Architecture of Auxiliary Network $A$}
On SNDCGAN CelebA-HQ with the optimal setting of $\sigma_\epsilon=0.5$, we experimented with different architectures for the auxiliary network $A$:
\begin{itemize}
    \item Linear Network: A linear network (with a single fully-connected layer) capacity was not sufficient to distinguish between generative transformations resulting from different $\epsilon$'s and hence the auxiliary task training failed.
    \noindent
    \vspace{-4pt}
    % and the accuracy of the self-supervision task remained close to $50\%$ throughout training.
    \item Non-linear Network: A non-linear network (with two fully-connected layers and ReLU activation at the hidden layer) achieved a FID score of $19.63$.
    % was able to achieve an accuracy of $>90\%$ on the self-supervision task by the end of training. It
    \noindent
    \vspace{-6pt}
    \item Convolutional Network: A convolutional network (with one convolutional layer, batch normalisation layer and fully-connected layer) achieved a FID score of $21.27$. 
\end{itemize}

\section{Qualitative Analysis of Generated Images}
\subsection{LT-BigGAN ImageNet}\label{sec:mode_collapse}

\begin{figure*}[t]
\centering
    \begin{minipage}{.48\textwidth}
    \includegraphics[width=\linewidth]{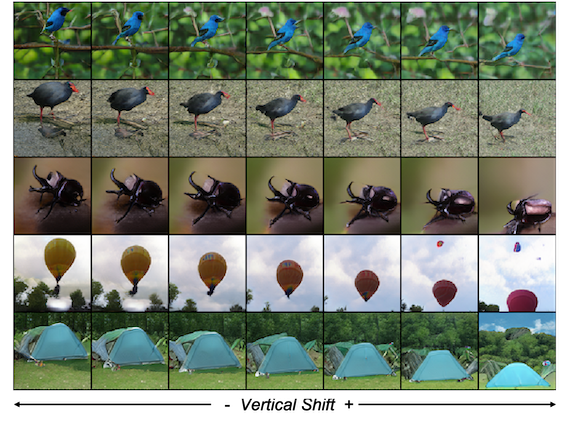}
    \end{minipage}%
    \hspace{0.05cm}
    \begin{minipage}{.48\textwidth}
    \includegraphics[width=\linewidth]{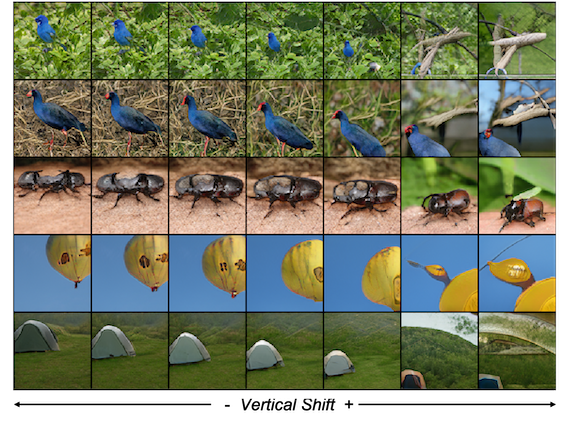}
    \end{minipage}
    \begin{minipage}{.48\textwidth}
    \includegraphics[width=\linewidth]{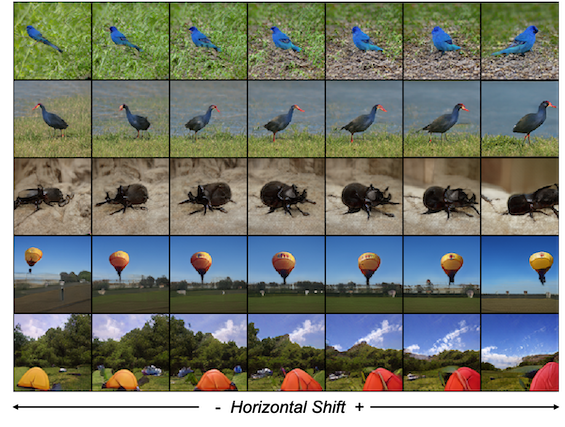}
    \subcaption{LT-BigGAN}
    \end{minipage}%
    \hspace{0.05cm}
    \begin{minipage}{.48\textwidth}
    \includegraphics[width=\linewidth]{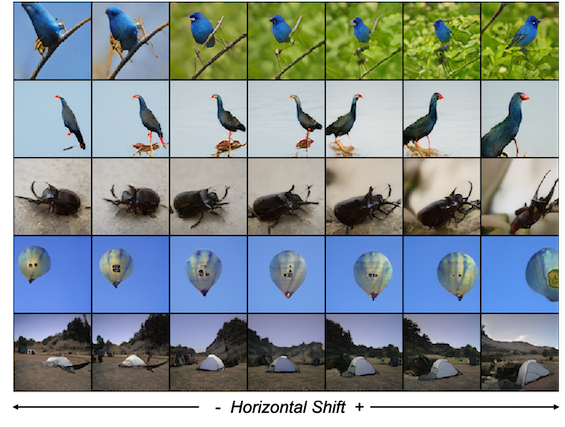}
    \subcaption{BigGAN}
    \end{minipage}
    \caption{\footnotesize{Qualitative comparison on geometric transformation horizontal and vertical shift between LT-BigGAN (left) and Baseline BigGAN (right) in five categories of ImageNet through latent space manipulation method of \cite{controlling2020iclr}}}
    \label{fig:lt-biggan_imagenet_horizontal_vertical}
\end{figure*}

\paragraph{Steerability of latent space}
We show more qualitative samples of varying the \textit{zoom}, \textit{brightness}, \textit{vertical position} and \textit{horizontal position} in generated images of classes same as \cite{controlling2020iclr} through latent space manipulation as discussed in Section 4.3 in the paper. Fig. \ref{fig:lt-biggan_imagenet_scale_zoom} shows sample images generated from LT-BigGAN and BigGAN model on perturbing the latent code in the positive and negative direction of brightness and zoom vector. Similarly, Fig. \ref{fig:lt-biggan_imagenet_horizontal_vertical} shows latent space steerability on horizontal and vertical shift. We can observe that the baseline model generates distorted images at the extremes and fails to control brightness factor and semantic content for all categories. In contrast, LT-BigGAN generates smooth variations of images while preserving the content and is able to generalize the brightness even for categories that usually are not available in a dark environment e.g \textit{cheeseburger} class.  

\paragraph{Mode Collapse} In the conditional image generation setting on ImageNet dataset using our proposed self-supervision approach, we observe that it not only improves the FID score but also helps in alleviating the issue of mode collapse. In Fig. \ref{fig:lt-biggan_imagenet_mode} we show example images of classes which suffer from mode collapse in a baseline BigGAN model trained on ImageNet and its corresponding samples generated from LT-BigGAN. We can see that images generated from LT-BigGAN are more diverse as compared to the baseline model.

\subsection{Image Editing on LT-StyleGAN CelebA-HQ }\label{sec:stylegan}
\begin{figure*}[t]
\centering
    \begin{minipage}{.48\textwidth}
    \includegraphics[width=\linewidth]{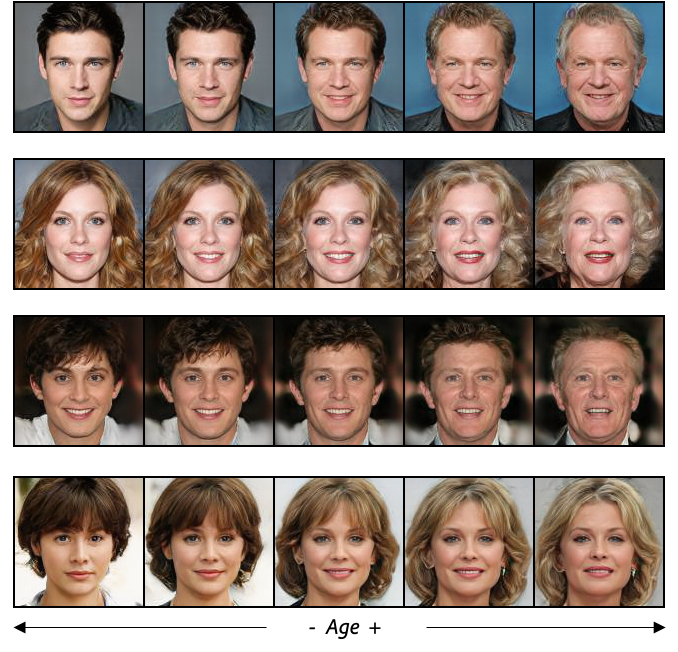}
    \end{minipage}%
    \hspace{0.05cm}
    \begin{minipage}{.48\textwidth}
    \includegraphics[width=\linewidth]{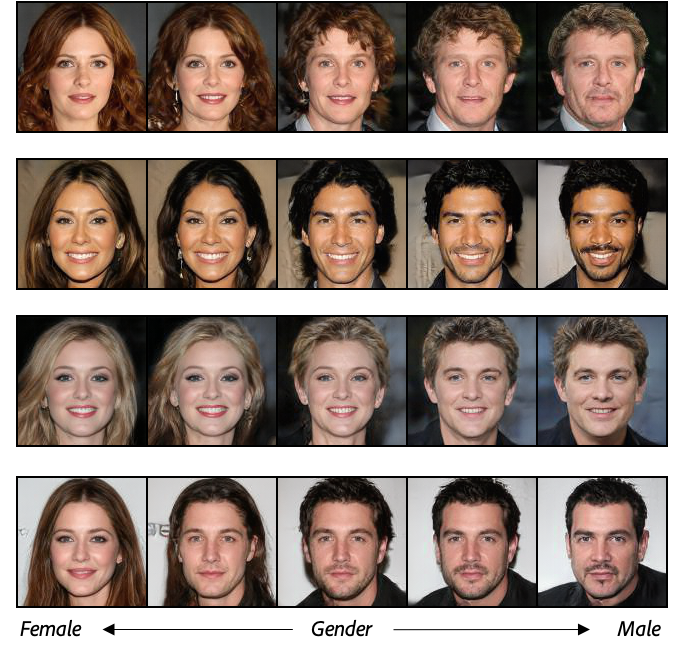}
    \end{minipage}\\
    \begin{minipage}{.48\textwidth}
    \includegraphics[width=\linewidth]{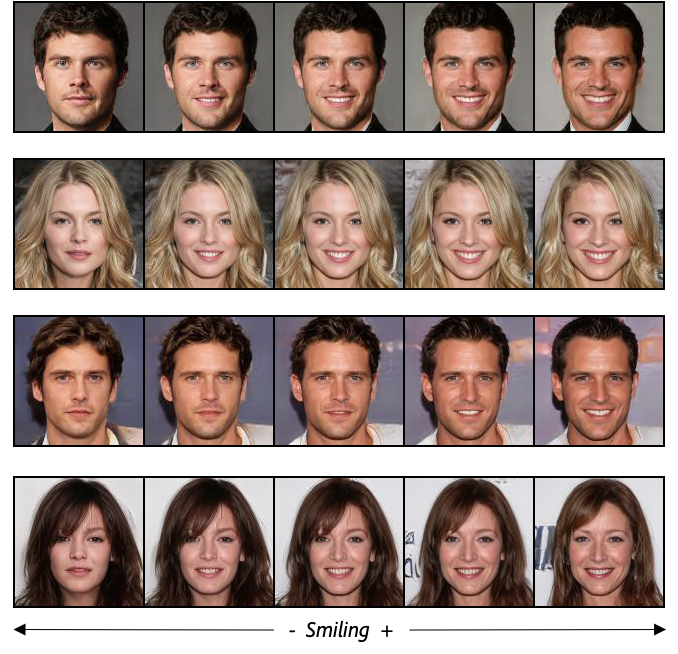}
    \end{minipage}%
    \hspace{0.05cm}
    \begin{minipage}{.48\textwidth}
    \includegraphics[width=\linewidth]{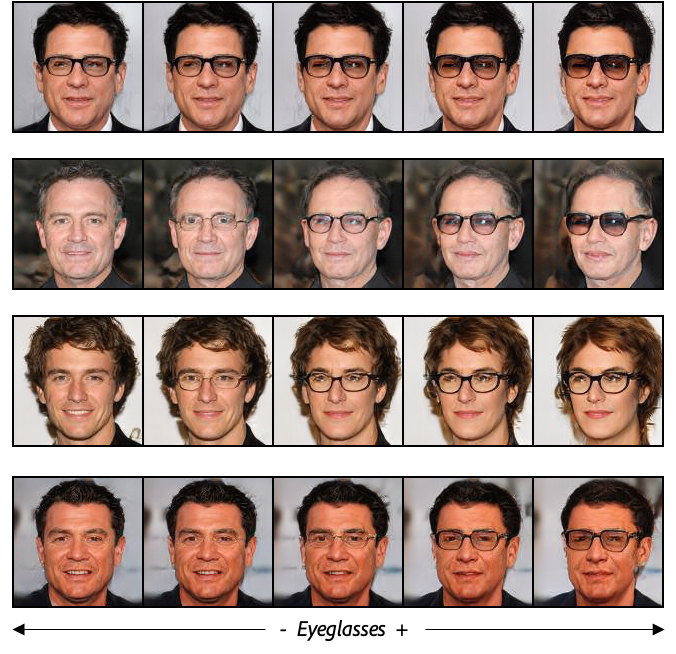}
    \end{minipage}
    \caption{\footnotesize{Manipulation of Age(top left), Gender(top right), Smile(bottom left)and Eyeglasses(bottom right) attributes by navigating the latent space of LT-StyleGAN using InterfaceGAN \cite{interface2020shen} framework. Original images are in the centre and the left and right images are generated by moving the latent code in negative and positive directions respectively.}}
    \label{fig:lt-stylegan2}
\end{figure*}

In Fig. \ref{fig:lt-stylegan2}, we show more examples of the manipulation of facial attributes namely age, gender, smile expression and eyeglasses by using the InterfaceGAN framework \cite{interface2020shen} on LT-StyleGAN model.

%%%%%%%%% BODY TEXT
\section{Hyper-parameter Details}\label{sec:hyperparameter}
This section mentions the choice of hyper-parameters for training LT-GAN over different datasets and architectures.
% We begin by describing the GAN architecture in detail that were used for the experiments, which we could not include in the main paper owing to space constraints. 
For experiments in CR-GAN, as mentioned in \cite{crgan2019chen}, the augmentation used for consistency regularization is a combination of randomly shifting the image by a few pixels and random horiozontal flipping. The shift size is 4 pixels for both CIFAR-10 and CelebA-HQ datasets and rest all hyper-parameters remain same as baseline. For our LT-GAN approach, we used twice the batch size for $G$ and kept the batch size of $D$ same as that of the baseline, because this modification achieved better results.  Note that for fair comparison, we also tried doubling the batch size of $G$ for baseline models, however the FID performance deteriorated.

% Some of the common hyper-parameter choices and terminologies used in this section are mentioned below:
% \begin{itemize}
%   \item SNDCGAN discriminator consists of 7 convolutional layers followed by a linear layer at the end. Each convolutional layer is followed by a ReLU activation. We treat each convolutional layer followed by its ReLU activation as a single layer. Thus, SNDCGAN discriminator consists of 8 layers.
%   \item As mentioned in CR-GAN~\cite{crgan2019chen}, the augmentation used for consistency regularization is a combination of randomly shifting the image by a few pixels and randomly flipping the image horizontally. The shift size is 4 pixels for both CIFAR-10 and CelebA-HQ datasets.
% \end{itemize}

\subsection{SNDCGAN CIFAR-10} 
Following the hyper-parameter choices of \cite{crgan2019chen}, we use $d_{step} = 1$ and set the dimensionality of the latent space to be $128$. Adam optimizer with $\alpha=0.0002$, $\beta_1=0.5$ and $\beta_2=0.999$ is used for both $G$ and $D$. We use a batch size of $64$ for both $G$ and $D$ for Baseline models. 

\textbf{LT-GAN:}
% We use batch sizes of $128$ and $64$ respectively for $G$ and $D$. 
$\sigma_\epsilon$ is chosen to be $0.6$ and Adam optimizer with default values of $\alpha=0.001$, $\beta_1=0.9$ and $\beta_2=0.999$ is chosen for the auxiliary network $A$. $\lambda$ is set to $1.0$. The encoder features $E(G(z))$ corresponding to generated images $G(z)$ are taken from the fifth layer of the discriminator \footnote{SNDCGAN discriminator consists of 7 convolutional layers followed by a linear layer at the end. Each convolutional layer is followed by a ReLU activation. We treat each convolutional layer followed by its ReLU activation as a single layer. Thus, SNDCGAN discriminator consists of 8 layers.}. The features $E(G(z))$ are passed through an average pool 2D layer with kernel size $2$, stride $2$ and zero padding, and then flattened before being passed to the auxiliary network. The number of warmup iterations $n$ before introducing the self supervision task is $2000$. 
\\

\textbf{CR+LT-GAN}:
% We use batch sizes of $128$ and $64$ respectively for $G$ and $D$. 
$\sigma_\epsilon$ is chosen to be $0.55$. Rest all hyper-parameters are same as those mentioned for LT-GAN.
% We use batch sizes of $128$ and $64$ respectively for $G$ and $D$. $\sigma_\epsilon$ is chosen to be $0.55$. Adam optimizer with default values of $\alpha=0.001$, $\beta_1=0.9$ and $\beta_2=0.999$ is chosen for the auxiliary network $A$. $\lambda$ is set to $1.0$. The encoder features $E(G(z))$ corresponding to generated images $G(z)$ are taken from the fifth layer of the discriminator. The features $E(G(z))$ are passed through an average pool 2D layer with kernel size $2$, stride $2$ and zero padding, and then flattened before being passed to the auxiliary network. The number of warmup iterations $n$ before introducing the self supervision task is chosen to be $1500$.

%%%%%%%%%%%%%%%%%%%%%%%%%%%%%%%%%%%%%%%%%%%%%%%%%%%%%%%%%%%%%%%%%%%%%%%%%%%%%%%%
\subsection{BigGAN CIFAR-10}
We use the standard value\cite
{biggan2018brock} of $d_{steps} = 4$, dimensionality of $z$ as $128$ and batch size of $64$. Adam optimizer with $\alpha=0.0002$, $\beta_1=0.0$ and $\beta_2=0.999$ is used for both $G$ and $D$.\\
% For LT-GAN we doubled the batch size for $G$ as explained in the above section.  
\vspace{-2pt}
\indent \textbf{LT-GAN:}
% We use batch sizes of $128$ and $64$ respectively for $G$ and $D$. 
$\sigma_\epsilon$ is chosen to be $0.6$. Adam optimizer with default values of $\alpha=0.001$, $\beta_1=0.9$ and $\beta_2=0.999$ is chosen for the auxiliary network $A$. $\lambda$ is set to $1.0$. The encoder features $E(G(z))$ corresponding to generated images $G(z)$ are taken from the last layer of the discriminator just before sum pooling. The number of warmup iterations $n$ before introducing the self supervision task is $2000$. 
\\
% \textbf{CR-GAN:}
% We use a batch size of $64$ for both $G$ and $D$. 
% \\
\indent \textbf{CR+LT-GAN}: All hyper-parameters are same as that of LT-GAN with default CR-GAN configuration.
% We use batch sizes of $128$ and $64$ respectively for $G$ and $D$. $\sigma_\epsilon$ is chosen to be $0.6$. The number of warmup iterations $n$ before introducing the self supervision task is chosen to be $2000$. Rest all hyper-parameters are same as those mentioned for LT-GAN.

%%%%%%%%%%%%%%%%%%%%%%%%%%%%%%%%%%%%%%%%%%%%%%%%%%%%%%%%%%%%%%%%%%%%%%%%%%%%%%%%
\subsection{SNDCGAN CelebA-HQ} 
Following the hyper-parameter choices of \cite{crgan2019chen}, we use $d_{steps} = 1$ and set the dimensionality of the latent space to be $128$. Adam optimizer with $\alpha=0.0002$, $\beta_1=0.5$ and $\beta_2=0.999$ is used for both $G$ and $D$. We use a batch size of $ 64$ for both $G$ and $D$ for baseline model.

% For our LT-GANs approach, we doubled the batch size for $G$ and kept the batch $D$ same as the baseline, because this modification achieved better results.  Note that for fair comparison, we also tried doubling the batch size of $G$ for baseline models, however, the FID performance was decreased.

% \textbf{Baseline:}
% We use a batch size of $ 64$ for both $G$ and $D$. \\
% \\
\textbf{LT-GAN:}
% We use batch sizes of $112$ and $64$ respectively for $G$ and $D$.
$\sigma_\epsilon$ is chosen to be $0.5$. Adam optimizer with default values of $\alpha=0.001$, $\beta_1=0.9$ and $\beta_2=0.999$ is chosen for the auxiliary network $A$. $\lambda$ is set to $1.0$. The encoder features $E(G(z))$ corresponding to generated images $G(z)$ are taken from the seventh layer of the discriminator. The features $E(G(z))$ are passed through an average pool 2D layer with kernel size $4$, stride $4$ and zero padding, and then flattened before being passed to the auxiliary network. The number of warmup iterations $n$ before introducing the self supervision task is chosen to be $1500$. \\

% \textbf{CR-GAN:}
% We use a batch size of $64$ for both $G$ and $D$. \\
% \\
\vspace{-8pt}
\textbf{CR+LT-GAN}:
% We use batch sizes of $100$ and $64$ respectively for $G$ and $D$.
The number of warmup iterations $n$ before introducing the self supervision task is chosen to be $5000$. Rest all hyper-parameters are same as those mentioned for LT-GAN.
% We use batch sizes of $100$ and $64$ respectively for $G$ and $D$. $\sigma_\epsilon$ is chosen to be $0.5$. Adam optimizer with default values of $\alpha=0.001$, $\beta_1=0.9$ and $\beta_2=0.999$ is chosen for the auxiliary network $A$. $\lambda$ is set to $1.0$. The encoder features $E(G(z))$ corresponding to generated images $G(z)$ are taken from the seventh layer of the discriminator. The features $E(G(z))$ are passed through an average pool 2D layer with kernel size $4$, stride $4$ and zero padding, and then flattened before being passed to the auxiliary network. The number of warmup iterations $n$ before introducing the self supervision task is chosen to be $5000$. As specified in \cite{crgan2019chen}, the augmentation used for consistency regularization is a combination of randomly shifting the image by $4.0$ pixels and randomly flipping the image horizontally. The consistency regularization coefficient is set to $10.0$.

%%%%%%%%%%%%%%%%%%%%%%%%%%%%%%%%%%%%%%%%%%%%%%%%%%%%%%%%%%%%%%%%%%%%%%%%%%%%%%%%
\subsection{StyleGAN CelebA-HQ}

StyleGAN adopts progressive growing of both the generator and the  discriminator networks. In LT-StyleGAN, we introduce the self supervision task after the layer corresponding to 128 resolution has completely faded into the network architecture. The per-pixel noise added after after each convolution block in generator is kept same while generating images corresponding to latent codes $z$ and $z+\epsilon$. For incorporating mixing regularization in LT-StyleGAN,the GAN-induced transformation of $G(z_1, z_2)$ is generated as $G(z_1+\epsilon_1, z_2+\epsilon_2)$, where $\epsilon_1$ and $+\epsilon_2$ are distinct.
% The networks start with low-resolution images, and then progressively increase the resolution by adding new layers. The addition of each new layer involves two phases: i) gradually fade in the new layer and ii) stabilise the network architecture before introducing the next layer. 
% wherein during training, a given percentage of images are generated using two latent codes instead of one. Such an image is generated by switching from one latent code to another at a randomly selected point in the synthesis network of the generator. Let us denote such an image as $G(z_1, z_2)$. Then, in LT-StyleGAN, the GAN-induced transformation of $G(z_1, z_2)$ is generated as $G(z_1+\epsilon_1, z_2+\epsilon_2)$, where $\epsilon_1$ and $+\epsilon_2$ are distinct.
  
Following the hyper-parameter choices of \cite{stylegan2019karras}, we set the dimensionality of both the latent spaces $Z$ and $W$ to be 512. The mapping network from $Z$ to $W$ is a 8 layer MLP. While training using progressive growing, we start from $8\times8$ resolution, fade in a new layer during the next 600K images and then let the network stabilize for next 600K images before introducing a new layer. For $128\times128$ resolution, we use Adam optimizer with $\alpha=0.0015$, $\beta_1=0.0$ and $\beta_2=0.99$ for both $G$'s synthesis network and $D$. We reduce the learning rate by two orders of magnitude for $G$'s mapping network (i.e. a learning rate of $\alpha=0.000015$), as specified in \cite{stylegan2019karras}. We use $d_{steps}=1$. We use a batch size of $32$ for both $G$ and $D$ with mixing probability to $0.9$ for baseline model.

% \\
% \\
% \textbf{Baseline:}
% For $128\times128$ resolution, we use a batch size of $32$ for both $G$ and $D$. We set mixing probability to $0.9$. \\
% \\
\textbf{LT-GAN:}
% For $128\times128$ resolution, we use batch size of $64$ and $32$ respectively for $G$ and $D$.
We choose $\sigma_\epsilon$ to be $0.5$ with a mixing probability of $0.5$. Adam optimizer with default values of $\alpha=0.001$, $\beta_1=0.9$ and $\beta_2=0.999$ is set for the auxiliary network $A$. We use $\lambda$ value of $0.5$. We take the encoder features $E(G(z))$ corresponding to generated images $G(z)$ from the layer of the discriminator corresponding to $16\times16$ resolution. The features $E(G(z))$ pass through an average pool 2D layer with kernel size $2$, stride $2$ and zero padding, and then we flatten it before passing it to the auxiliary network.

\subsection{BigGAN ImageNet}
We use the default configuration$^4$ of $d_{step} = 1$, dimensionality of $z$ as $120$, batch size of $8*256$. We select Adam optimizer with $\alpha=0.0001$ and $0.0004$ for $G$ for $D$ respectively. 

\textbf{LT-GAN:}
% We use batch sizes of $112$ and $64$ respectively for $G$ and $D$.
We choose $\sigma_\epsilon$ to be $0.5$. We set Adam optimizer with default values of $\alpha=0.001$, $\beta_1=0.9$ and $\beta_2=0.999$ for the auxiliary network $A$. $\lambda$ is set to $0.5$. We take the encoder features $E(G(z))$ corresponding to generated images $G(z)$ from the seventh layer of the discriminator. The features $E(G(z))$ pass through an average pool 2D layer with kernel size $2$, stride $2$ and zero padding, and then we flatten before passing to the auxiliary network. The number of warmup iterations $n$ before introducing the self supervision task is set to $100K$ \footnote{We use the pretrained model of \url{https://github.com/ajbrock/BigGAN-PyTorch} as baseline and use the provided checkpoint at $100K$ for fine-tuning with LT-GAN}.

\end{document}